\documentclass[conference,10pt]{IEEEtran}

\usepackage{microtype}
\usepackage{graphicx}
\usepackage{subfigure}
\usepackage{booktabs}
\usepackage{array}
\usepackage{amsmath,amssymb}
\usepackage{multirow}
\usepackage[table]{xcolor}
\usepackage{placeins}
\usepackage{hyperref}
\usepackage{etoolbox}
\usepackage{adjustbox}
\usepackage{etoolbox}
\usepackage{adjustbox}

\definecolor{arc}{RGB}{0,0,0}
\definecolor{vlc}{RGB}{230,159,0}
\definecolor{ssc}{RGB}{0,114,178}
\definecolor{vvc}{RGB}{213,94,0}

\newcolumntype{G}[1]{>{\columncolor{vlc!#1}}r}
\newcolumntype{A}[1]{>{\columncolor{arc!#1}}r}
\newcolumntype{V}[1]{>{\columncolor{vvc!#1}}r}
\newcolumntype{S}[1]{>{\columncolor{ssc!#1}}r}

\newcommand{\sys}{SpecSearch}

\begin{document}
\raggedbottom
\title{\sys: Towards Hardware-Aware Efficient Speculative Decoding}
\author{
\IEEEauthorblockN{Selin Yildirim, Ali Zeyrek and Deming Chen}
\IEEEauthorblockA{
University of Illinois Urbana-Champaign, IL, USA\\
\{seliny2, azeyrek2, dchen\}@illinois.edu
}
}
\maketitle

\begin{abstract}
Speculative decoding (SD) accelerates large language model (LLM) inference, but its optimal configuration depends jointly on workload characteristics, GPU hardware, and available memory. For example, on the RTX~5090, the state-of-the-art speculative drafter DFlash achieves $3.50\times$ the throughput of autoregressive (AR) decoding at low concurrency, but only $0.76\times$ at high concurrency with short outputs. We present \sys{}, a hardware-aware search engine that jointly optimizes decoding strategy, draft length, and precision across workload--GPU pairs. Its analytical cost model captures memory and compute constraints to identify feasible configurations and narrow the search space for throughput optimization. Guided by this model, calibration measurements are collected from multiple GPUs. Using the measurements, a Jev-based predictor learns to select high-throughput decoding configurations on held-out GPUs without requiring deployment-GPU calibration. The predictor is parameter-efficiently trained on calibration data, while adapting a frozen Qwen3-4B using LoRA adapters and a linear correction layer. Across NVIDIA RTX~5090, H100, L40 and AMD MI300X GPUs, \sys{} with full calibration achieves an average of $2.01\times$  the throughput of AR and $1.36\times$ the throughput of SD, both served by vLLM. The proposed Jev-based predictor closely matches this search performance with $1.30\times$ throughput improvement over SD with vLLM, while eliminating the need for held-out GPU calibration. Additionally, it retains $97.36\%$ of the throughput achieved by \sys{} that utilizes full calibration.
\end{abstract}

\begin{figure*}[t]\centering
  \includegraphics[width=0.9\textwidth]{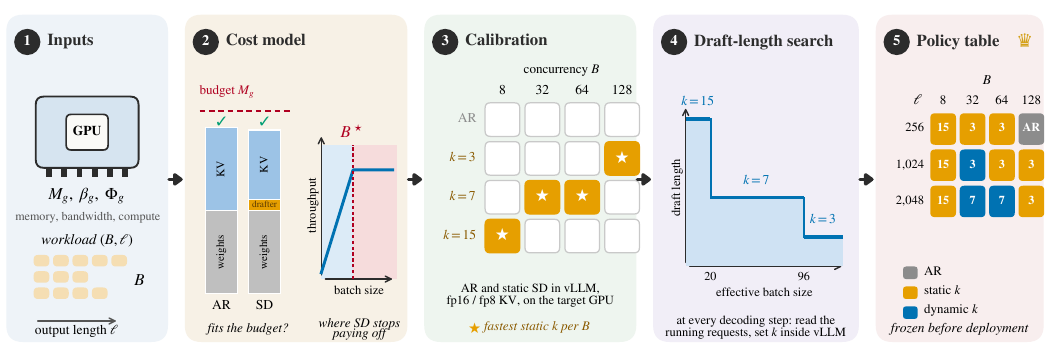}
  \caption{\textbf{Overview of \sys{} with full calibration :} Given the GPU and the workload (1), the proposed cost model discards deployment configurations that exceed the
  memory budget and determines the critical batch size $B^\star$ of the device (2). Calibration on the deployment GPU runs AR and SD where quantization and optimal draft lengths $k$ are explored within vLLM (3). \sys{} explores dynamic draft-length mapping for vLLM that sets $k$ from the effective batch size at every decoding step (4). The most profitable configuration for each workload is frozen into a
  policy lookup table to be reused during real-time inference (5).}\label{fig:overview}
  \end{figure*}
\section{Introduction}
Speculative decoding (SD)~\cite{leviathan2023fast,chen2023accelerating} reduces autoregressive (AR) decoding latency: an inexpensive drafter model proposes several draft tokens, and the target model verifies them in parallel in a single forward pass. Verification makes generation lossless, so the output distribution equals that of the target model. State-of-the-art diffusion-LLM drafters such as DFlash~\cite{chen2026dflash} propose a whole block of draft tokens in a single drafter forward pass. Although advances in drafter architectures are rapidly improving speculative decoding efficiency, the achievable speedup remains highly dependent on deployment-time decisions: whether to load a drafter model at all, how many tokens to draft (i.e., the draft length) and at which precision to store the model weights and the key--value (KV) cache. These decisions are particularly critical on memory-constrained GPUs. Model weights and speculation buffers consume a major portion of device memory. Consequently, the memory left for the KV cache determines how many requests the inference engine can serve concurrently. This limitation hinders efficient speculative decoding on memory-constrained GPUs.

\begin{figure*}[t]\centering
\includegraphics[width=0.85\textwidth]{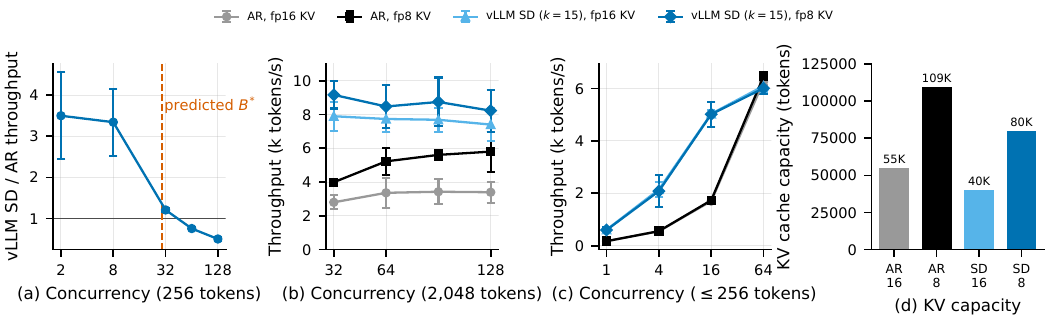}
\caption{The best decoding configuration and its memory cost depend on the workload. (a) vLLM SD-to-AR throughput ratio with 256-token outputs and fp16 KV; the dashed line marks the critical batch size $B^\star$ predicted by \sys{} cost model (Section~\ref{sec:method}). (b) With longer outputs (2,048 tokens), vLLM SD with fp8 KV is fastest from concurrency $32$ to $128$. (c) With shorter outputs (up to 256 tokens), vLLM SD has $3.5$--$3.9\times$ throughput at concurrency $1$--$4$, but its advantage over AR disappears by concurrency $64$. (d) The drafter use reduces KV cache capacity to $0.73\times$ that of AR at equal KV precision.}\label{fig:port}
\end{figure*}

Modern inference engines such as vLLM~\cite{kwon2023vllm} run kernels optimized for the deployment hardware. They further enhance aggregate throughput with continuous batching, which admits waiting requests into the running batch as others finish. Additionally, they reduce memory footprint through paging KV cache. Nonetheless, they leave the algorithmic configuration of SD to the user: the draft length and the decoding precision are manual knobs, typically set once and reused across workloads and hardware. We observe that this practice loses substantial throughput, because \emph{the best decoding configuration, including whether to speculate at all, depends jointly on the workload and the hardware}. Fig.~\ref{fig:port} illustrates this for Qwen3-4B with a DFlash drafter on a resource-constrained GPU. In Fig.~\ref{fig:port}(a), with shorter text generation, vLLM SD is $3.50\times$ faster than AR decoding at 2 concurrent requests but only $0.76\times$ as fast at 64, because verification becomes compute-bound. In Fig.~\ref{fig:port}(b), with longer text generation, vLLM SD with an fp8 KV cache is the fastest configuration at every concurrency from 32 to 128, ahead of AR decoding and of vLLM SD with an fp16 KV cache. The concurrency at which speculation stops paying off therefore depends on the output length, the memory budget, and the ridge point of the device, i.e., the ratio of its peak compute throughput to its memory bandwidth.

We present \sys{}, which automatically identifies the best-performing decoding configuration for a given GPU and workload in terms of throughput and wall-clock speedup. A decoding configuration comprises a set of deployment policies, including whether to enable speculation, setting a static or dynamic draft length based on effective batch size (concurrency), as well as the precision of model weights and the KV cache. We characterize each workload by the number of concurrent requests ($B$), which we refer to interchangeably as concurrency or batch size, and their generation lengths. \sys{} has two components:

\textbf{First}, an analytical cost model characterizes the device's memory budget, memory bandwidth, and peak compute throughput. It determines whether a deployment configuration fits within the available memory and, if feasible, estimates the maximum concurrency $(B)$ that the KV cache can support without request preemption. It also identifies the critical batch size $(B^\star)$, beyond which compute contention diminishes the throughput benefits of speculation.

\textbf{Second}, \sys{} selects the most profitable decoding configuration for each workload on a given GPU through two complementary approaches. In \textbf{full calibration mode}, \sys{} uses calibration measurements guided by the cost model to search the policy space and identify the best-performing configuration for each workload. These configurations are then stored in a lookup table before real-time inference deployment. This workflow is illustrated in Figure \ref{fig:overview}. We further use the measured calibration data to construct a training dataset for fine-tuning a \emph{Jev-based policy selector}~\cite{li2026llmasjev}. Alternatively, in \textbf{learning-based mode}, \sys{} leverages this trained model to predict the optimal decoding configuration for each workload on a given GPU, eliminating the need for prior calibration on held-out GPUs.

\textbf{Contributions.}
\begin{enumerate}
\item \emph{Defining a joint optimization space}: We provide evidence that the optimal speculative decoding (SD) configuration depends both on the workload and the deployment GPU (Section \ref{sec:motiv}). \sys{} introduces a hardware-aware cost model to narrow the search space for finding performance-optimal decoding policies.
\item \emph{Validation and portability of cost model}: \sys{}'s cost model estimates total GPU memory footprint for LLM inference within $0.1\%$ and $0.2\%$ on the AMD MI300X~\cite{amd2023mi300x} and NVIDIA RTX~5090~\cite{nvidia2025rtx5090} GPUs, respectively. Furthermore, it estimates a maximum serviceable concurrency within $0.5\%$ and $0.6\%$ on the GPUs, respectively. The cost model also correctly predicts whether a workload fits a device budget under all evaluated configurations on the two GPUs.
\item \emph{End-to-end performance gains}: \sys{} with full calibration achieves $2.16\times$ and $1.68\times$ the throughput of AR on RTX~5090 and MI300X, respectively, averaged across several workloads. Furthermore, it achieves $1.44\times$ and $1.21\times$ speedup over baseline SD served by vLLM, respectively. Across four different GPUs, it delivers an average $1.36\times$ speedup over vLLM SD. The Jev-based predictor closely approaches the full calibration \sys{} performance, retaining $97.36\%$ of its throughput and achieving $1.30\times$ the throughput of vLLM SD.
\end{enumerate}

\section{Background and Related Work}\label{sec:related}
\textbf{Speculative decoding.} Draft-then-verify approach was introduced with an independent draft model~\cite{leviathan2023fast,chen2023accelerating}. Medusa uses parallel decoding heads with tree attention~\cite{cai2024medusa}. EAGLE drafts at the level of target features~\cite{li2024eagle}. EAGLE-3 drafts tokens from fused multi-layer target features~\cite{li2025eagle3}. State-of-the-art (SoTA) DFlash drafters propose a block of tokens in a single forward pass of a block-diffusion model conditioned on the target features~\cite{chen2026dflash}. All methods deliver lossless generation with respect to the target distribution. We use DFlash drafters for evaluating SD throughout.

\textbf{Paged KV cache and memory-efficient speculation.} vLLM~\cite{kwon2023vllm} and SGLang~\cite{zheng2024sglang} are LLM serving frameworks that employ paged KV-cache management to support efficient memory allocation and high-throughput serving. vLLM~\cite{kwon2023vllm} introduces PagedAttention, which manages KV-cache entries in non-contiguous, fixed-size blocks, reducing memory fragmentation and enabling flexible KV-cache sharing and allocation across requests. Building on that, \sys{} proposes a cost model that estimates the maximum serviceable concurrency a paged KV cache can support. Prior works on memory-efficient speculative decoding include SpecMemo~\cite{yildirim2025specmemo}, which optimizes runtime buffer allocation in Medusa~\cite{cai2024medusa} to improve single-request throughput and reduce memory footprint. However, it does not account for concurrency under paged KV-cache management or workload-dependent throughput.

\textbf{Dynamic draft-length selection.} vLLM and SGLang expose configurable speculative decoding parameters, including the draft length, which is the number of consecutive tokens proposed by the drafter. vLLM supports user-defined mappings from batch-size ranges to draft lengths. SGLang provides adaptive SD that adjusts draft length based on observed acceptance rates within a workload. However, neither determines an optimal draft length based on both the workload and deployment hardware. \sys{} automates this process through GPU-specific calibration, selecting draft-length policies based on measured throughput and which precision to use. Thus, \sys{} performs resource- and workload-aware dynamic draft length selection.

\textbf{Quantization.} Model-weight ~\cite{frantar2023gptq,xiao2023smoothquant,lin2024awq} and KV-cache quantization~\cite{liu2024kivi,hooper2024kvquant,he2024zipcache,li2025kvtuner} are established techniques for reducing the memory footprint of LLM inference. \sys{} incorporates both model-weight precision and KV-cache precision as configurable dimensions of speculative decoding (SD), and quantifies their individual and combined effects on end-to-end performance.

\textbf{Adapting speculation to workload and hardware.} Existing works address complementary aspects of optimizing SD. Sequoia profiles hardware-dependent draft and verification costs to select speculative tree size and depth~\cite{chen2024sequoia}. MagicDec~\cite{sadhukhan2024magicdec} models the transition from memory-bound to compute-bound execution as batch size increases in self-speculative decoding, and uses this analysis to optimize speculation for long-context workloads. At the serving layer, TurboSpec controls draft lengths under changing load~\cite{liu2024turbospec}. AdaServe constructs hardware-aware speculative decoding trees for per-request latency objectives~\cite{li2026adaserve}. Unlike prior work, \sys{} addresses a broader deployment-level choice: whether to use AR or SD, which static or dynamic draft length to use, and which precision fits the deployment GPU's resource constraints. The proposed cost model analyzes workloads and calibrates them on GPUs to select a workload- and GPU-specific decoding policy before serving.

\textbf{Learning-based deployment optimization.} Prior work predicts performance of deep learning workloads across hardware platforms to reduce the cost of hardware profiling. Habitat scales operation-level measurements from a source GPU to estimate training iteration time on other GPUs~\cite{yu2021habitat}. NeuSight predicts training and inference latency on GPUs using learned tile-level models constrained by hardware performance bounds~\cite{lee2025neusight}. For LLM inference, LLMCompass models hardware architectures and explores operator mappings and schedules to estimate latency and analyze performance--cost tradeoffs~\cite{zhang2024llmcompass}. These works provide a foundation for making deployment decisions across GPUs. Building on this direction, \sys{} proposes training a policy selector on GPU performance measurements to select optimal decoding configurations. Recent Jev~\cite{li2026llmasjev} is a lightweight decision model that predicts choices among predefined alternatives, enabling efficient learned decision-making. Unlike prior approaches, \sys{} fine-tunes a Jev-based policy selector for predicting the optimal decoding configuration across LLM workloads and hardware.

\section{Motivation}\label{sec:motiv}

We preliminarily evaluate AR and SD using vLLM with FP16 and FP8 KV-cache precisions on an RTX 5090. We find that the best-performing configuration depends jointly on the workload and GPU.

\textbf{Observation 1: The concurrency at which speculation loses its advantage depends on the workload and GPU.} As concurrency increases, decoding shifts from memory-bound to compute-bound execution~\cite{williams2009roofline}, reducing speculation's benefit. The transition depends on the GPU's compute-to-bandwidth ratio, which our cost model estimates through the critical batch size $B^\star$ (Table~\ref{tab:dev}). SD throughput advantage for 256-token outputs drops from $3.50\times$ at concurrency 2 to $1.21\times$ at 32, then reverses to $0.76\times$ at 64 (Fig.~\ref{fig:port}(a)). We validate this finding on AMD MI300X GPU, where the SD gain is $1.13\times$ at concurrency 32. For 2,048-token outputs, SD remains advantageous beyond concurrency 128 (Fig.~\ref{fig:port}(b)).

\textbf{Observation 2: KV quantization benefits long outputs.} For 4,096-token outputs at concurrency 48, throughput reaches 8,479 tokens/s with SD and fp8 KV, versus 6,813 with SD and fp16 KV. This yields a $1.24\times$ speedup, with nearly identical acceptance lengths (11.8 vs. 11.9). The benefit persists for 2,048-token outputs: fp8 KV improves SD throughput by $1.10$--$1.16\times$ across concurrency levels 32--128 (Fig.~\ref{fig:port}(b)). For outputs up to 256 tokens, however, the difference is at most 4\% (Fig.~\ref{fig:port}(c)). Thus, KV quantization primarily benefits long-output workloads where KV-cache capacity limits concurrency. 

\textbf{Observation 3: Speculation reduces KV-cache capacity.} Drafter weights and runtime buffers consume memory otherwise available for KV caching. Under a 15.7\,GiB budget on RTX~5090, vLLM SD retains only $0.73\times$ the AR KV-cache capacity at equal KV precision (Fig.~\ref{fig:port}(d)).

These observations motivate a joint search mechanism for optimal decoding policy selection across workloads and GPUs. \sys{} combines an analytical cost model with device calibration to capture hardware-specific costs at run-time.

\begin{table}[!htbp]\centering\footnotesize
\caption{Device parameters of \sys{} cost model: memory capacity, peak theoretical bandwidth $\beta_g$, and peak dense fp16 matrix throughput $\Phi_g$~\cite{amd2023mi300x,nvidia2025rtx5090}. $B^\star$ denotes the GPU roofline knee point.}\label{tab:dev}
\setlength{\tabcolsep}{3pt}
\begin{tabular}{lrrrr}
\toprule
GPU & GB & $\beta_g$ (TB/s) & $\Phi_g$ (TF/s) & predicted $B^\star$\\
\midrule
MI300X & 192 & 5.30 & 1{,}307 & 30.8\\
RTX~5090 & 32 & 1.79 & 209.5 & 29.3\\
\bottomrule
\end{tabular}
\end{table}

\section{\sys{}}\label{sec:method}

\sys{} searches for an optimal deployment policy for LLM inference that maximizes throughput and speedup on a given GPU. In \textbf{full calibration mode}, \sys{} uses the cost model to define the policy search space and performs calibration runs to evaluate the policies. In \textbf{learning-based mode}, a fine-tuned Jev-based model infers the optimal policy on held-out GPUs without calibration.

\textbf{Notation.} GPU $g$ has nominal memory bandwidth $\beta_g$, peak fp16 compute throughput $\Phi_g$ (Table~\ref{tab:dev}), and inference memory budget $M_g$. The target and drafter have $P_t,P_d$ parameters and $L_t,L_d$ layers. We denote the number of KV heads by $H_{kv}$, head dimension by $d$, hidden size by $h$, and vocabulary size by $V$. A policy $\theta$ specifies draft length $k$, target and drafter weight sizes $b_w^t,b_w^d\in\{2,1,0.5\}$ bytes (fp16, fp8, int4), KV element size $b_{kv}\in\{2,1\}$ bytes (fp16, fp8), and execution mode (CUDA/HIP graphs or eager). A workload $(B,\ell)$ specifies batch size $B$ and generation length $\ell$. Given prompt length $S_0$, the per-request context length is $S=S_0+\ell$.

\subsection{Cost Model}
\textbf{Memory model.} Peak memory is
\begin{multline}\label{eq:mem}
M_{\text{peak}}(\theta)=(P_t b_w^{t}+P_d b_w^{d})+\kappa SB+n_c h S_0 b\\
+2(k+1)V B b+C_{\text{graph}}+\rho_g,
\end{multline}
where the terms represent model weights, KV caches, the context buffer for $n_c$ target layers, verification logits, CUDA/HIP graph memory (zero in eager mode), and runtime/allocator overhead, respectively. Here, $b$ is the activation element size in bytes, and $\kappa=2(L_t+L_d)H_{kv}d\,b_{kv}$ is the KV-cache cost per context token. The drafter cache stores keys and values projected from target model hidden states.

\textbf{Concurrency model.} The available KV-cache capacity, measured in tokens, is
\begin{equation}\label{eq:cap}
N_{\text{tok}}(\theta)=
\frac{M_g-W(\theta)-C_{\text{act}}-C_{\text{graph}}-\rho_g}{\kappa},
\end{equation}
where $W(\theta)=P_t b_w^t+P_d b_w^d$ and $C_{\text{act}}$ includes the context buffer and verification logits. Maximum supported concurrency is
\begin{equation}
C_{\max}(\theta,\ell)=\frac{N_{\text{tok}}(\theta)}{S}.
\end{equation}
A configuration is feasible at required concurrency $C_{\text{req}}$ if $N_{\text{tok}}(\theta)\ge C_{\text{req}}S$.

\textbf{Throughput model.} Following the roofline model~\cite{williams2009roofline}, we approximate decode-step time as
\begin{equation}\label{eq:step}
t_{\text{step}}\approx
\max\left(
\frac{W(\theta)}{\beta_g},
\frac{B(k+1)\phi}{\Phi_g}
\right),
\end{equation}
where $\phi\approx2P_t$ is target arithmetic per token. The terms capture weight reads and computation over $B(k+1)$ tokens, respectively. At low concurrency, decoding is memory-bound, and speculation incurs approximately one AR verification step while accepting $\tau$ draft tokens per step, yielding an approximate speedup of $\bar\tau$. At high concurrency, decoding becomes compute-bound, reducing the approximate speedup to $\bar\tau/(k+1)$.

Equating the two terms for AR decoding ($W=P_t b_w^t$) gives the critical batch size
\begin{equation}\label{eq:ridge}
B^\star\approx
\frac{P_t b_w^t}{\beta_g}\frac{\eta\Phi_g}{\phi}
=\frac{\eta b_w^t}{2}\frac{\Phi_g}{\beta_g},
\end{equation}
where $\eta\le1$ is the fraction of peak compute throughput achieved during decoding. Model size cancels, making $B^\star$ primarily dependent on the device's compute-to-bandwidth ratio and target weight precision. We use $B^\star$ to estimate the transition from memory-bound to compute-bound decoding.

\textbf{Joint cost model.} Given quality tolerance $\varepsilon$, we select
\begin{align}
\theta^*(B,\ell)&=\arg\max_{\theta\in\Theta_g}
\operatorname{Tput}_g(\theta,B,\ell) \notag\\
&\text{s.t.}\quad M_{\text{peak}}(\theta)\le M_g,\quad
\operatorname{QGap}(\theta)\le\varepsilon.
\label{eq:select}
\end{align}
Here, $\Theta_g$ contains configurations feasible under the device memory budget. The quality constraint applies only to lossy target-weight and KV precisions; draft length and drafter precision preserve the target distribution. Since the analytical model cannot capture low-level kernel efficiency, the full calibration setting measures $\operatorname{Tput}_g$ on the deployment GPU. On held-out GPUs, our zero-shot approach ranks static candidates using predictions or selection scores derived from source-GPU measurements.

\begin{table*}[!tp]\centering\scriptsize
\caption{Validation of the cost model across policies. Predicted (pred.) and measured (meas.) denote cost-model predictions for memory footprint and realized allocation, respectively. Additionally, GPU memory budgets are emulated, ranging from 9 to 15.7 GiB, to determine deployment feasibility. Thus, each cell contains two symbols: the first is the cost model prediction for feasibility and the second is the vLLM deployment outcome. A \checkmark{} means feasible (runs successfully), $\times$ means out-of-memory (fails).}\label{tab:cost}
\setlength{\tabcolsep}{2pt}
\begin{tabular}{@{}l rrr rrr ccccc@{\hspace{8pt}}rrr rrr ccccc@{}}
\toprule
 & \multicolumn{11}{c}{MI300X} & \multicolumn{11}{c}{RTX~5090}\\
\cmidrule(lr){2-12}\cmidrule(l){13-23}
 & \multicolumn{3}{c}{Memory (GB)} & \multicolumn{3}{c}{$C_{\max}$} & \multicolumn{5}{c}{Starts within allocation (GiB)} & \multicolumn{3}{c}{Memory (GB)} & \multicolumn{3}{c}{$C_{\max}$} & \multicolumn{5}{c}{Starts within allocation (GiB)}\\
\cmidrule(lr){2-4}\cmidrule(lr){5-7}\cmidrule(lr){8-12}\cmidrule(lr){13-15}\cmidrule(lr){16-18}\cmidrule(l){19-23}
configuration & pred. & meas. & err. & pred. & meas. & err. & 9 & 10 & 11 & 12 & 15.7 & pred. & meas. & err. & pred. & meas. & err. & 9 & 10 & 11 & 12 & 15.7\\
\midrule
AR, fp16 KV & 17.01 & 17.02 & $+0.1\%$ & 11.07 & 11.05 & $-0.2\%$ & $\times$/$\times$ & \checkmark/\checkmark & \checkmark/\checkmark & \checkmark/\checkmark & \checkmark/\checkmark & 16.92 & 16.93 & $+0.1\%$ & 13.24 & 13.22 & $-0.2\%$ & \checkmark/\checkmark & \checkmark/\checkmark & \checkmark/\checkmark & \checkmark/\checkmark & \checkmark/\checkmark\\
AR, fp8 KV & 17.01 & 17.02 & $+0.1\%$ & 22.14 & 22.11 & $-0.1\%$ & $\times$/$\times$ & \checkmark/\checkmark & \checkmark/\checkmark & \checkmark/\checkmark & \checkmark/\checkmark & 16.89 & 16.92 & $+0.2\%$ & 26.48 & 26.33 & $-0.6\%$ & \checkmark/\checkmark & \checkmark/\checkmark & \checkmark/\checkmark & \checkmark/\checkmark & \checkmark/\checkmark\\
vLLM SD, fp16 KV & 17.54 & 17.56 & $+0.1\%$ & 7.14 & 7.10 & $-0.5\%$ & $\times$/$\times$ & $\times$/$\times$ & $\times$/$\times$ & \checkmark/\checkmark & \checkmark/\checkmark & 17.29 & 17.31 & $+0.1\%$ & 8.83 & 8.80 & $-0.4\%$ & $\times$/$\times$ & $\times$/$\times$ & \checkmark/\checkmark & \checkmark/\checkmark & \checkmark/\checkmark\\
vLLM SD, fp8 KV & 17.54 & 17.56 & $+0.1\%$ & 14.28 & 14.20 & $-0.5\%$ & $\times$/$\times$ & $\times$/$\times$ & \checkmark/\checkmark & \checkmark/\checkmark & \checkmark/\checkmark & 17.28 & 17.31 & $+0.2\%$ & 17.66 & 17.57 & $-0.5\%$ & $\times$/$\times$ & \checkmark/\checkmark & \checkmark/\checkmark & \checkmark/\checkmark & \checkmark/\checkmark\\
\bottomrule
\end{tabular}
\end{table*}

\subsection{\sys{} with full calibration}\label{sec:details}
The cost model identifies the set $\Theta_g$ of feasible deployment policies for a workload on a given GPU, which may include varying precision settings for model weights and KV cache, as well as a draft length mapping. \sys{} performs an offline calibration pass for each feasible policy determined by the cost model on the deployment GPU. Accordingly, it ranks these deployment policies by their mean wall-clock speedup and throughput relative to AR decoding. The best policy (whether SD or AR) is then stored in a lookup table for inference-time selection before deploying the workload.

The effective batch size, defined as the number of scheduled requests per decoding step, varies as requests finish, are admitted, or are preempted in vLLM. Together with KV-cache precision, it determines GPU resource contention and KV-cache pressure. Thus, the optimal draft length $k$ also varies with the effective batch size and precision. 

\textbf{Dynamic draft-length search.} \sys{} constructs a dynamic draft length mapping from calibration data, identifying the fastest-performing $k$ for each workload and precision. At each decoding step, \sys{} uses the mapping to dynamically set the corresponding draft length \(k\) within vLLM. Table \ref{tab:maps} provides an example of the dynamic draft length mapping derived by \sys{}. Batches 21–48 use \(k=3\) for $\ell=256$ and $\ell=1,024$, but \(k=7\) for $\ell=2,048$. With fp8 KV and $\ell=2,048$, the general trend is \(k=15\), 7, and 3, respectively, for batches 1–20, 21–48, and 49–128. 

\subsection{Learning-Based Policy Predictor}\label{sec:learning-method}

\sys{} trains a lightweight Jev-based policy predictor~\cite{li2026llmasjev} to eliminate calibration on held-out GPUs. Given a deployment GPU and workload $(B,\ell)$, the predictor selects the decoding mode, draft length, and KV-cache precision. 
    
\textbf{Training and validation datasets.} The dataset covers eight decoding configurations: AR and SD with $k\in\{3,7,15\}$, each using fp16 or fp8 KV cache. Each GPU runs 96 configurations across $B\in\{8,32,64,128\}$ and $\ell\in\{256,1024,2048\}$. We use four distinct GPUs when creating the dataset: two for training, one for validation, and one held out for evaluation. For each training GPU, the highest-throughput configuration per workload serves as the training label.

\textbf{Model architecture.} We adapt Qwen3-4B into an eight-choice Jev-style decision model by training rank-8 LoRA adapters on its attention projections. Each choice represents one of the eight decoding configurations in the training dataset. We additionally train a linear answer-logit correction layer using cross-entropy loss, while keeping the pretrained backbone frozen. 

\textbf{Training objective.} Each training example \(i\) represents one configuration-selection problem for a deployment GPU and workload \((B,\ell)\). Its
  structured prompt contains the GPU and workload metadata, eight candidate configurations labeled A--H in shuffled order, and the permitted source
  calibration records. The label \(y_i\) is the answer letter assigned to the candidate with the highest measured throughput for that GPU--workload pair.
  Given the prompt, Qwen produces a final hidden representation \(h_i\), from which we compute the eight answer logits as
  \[
  z_i = W_{\mathrm{A:H}}h_i + W_c h_i,
  \]
  where \(W_{\mathrm{A:H}}\) denotes the frozen rows of Qwen's language-model output head corresponding to tokens A--H, and \(W_c\) is a trainable,
  zero-initialized linear correction layer. Rank-8 LoRA adapters on the query, key, value, and output attention projections modify the hidden
  representation \(h_i\), while the pretrained Qwen weights remain frozen. We optimize the LoRA parameters and \(W_c\) using the classification
  cross-entropy objective
  \[
  \mathcal{L}=-\frac{1}{N}\sum_{i=1}^{N}
  \log\left(\operatorname{softmax}(z_i)_{y_i}\right).
  \]
  Thus, training encourages the model to assign the highest answer score to the measured throughput-maximizing configuration. Further details are provided in Appendix~\ref{app:transfer-training}.

\textbf{Learning-based policy prediction.} At deployment time, the trained Jev-based predictor takes a structured text input, termed a \emph{decision prompt}, describing the deployment GPU, workload, and eight candidate configurations labeled A through H. Rather than generating text, the model scores the corresponding answer-label tokens and selects the configuration with the highest score. This enables policy selection on a held-out GPU without calibration or profiling, leveraging the relationships between workloads, hardware characteristics, and decoding configurations learned during training.

\section{Experimental Setup}\label{sec:setup}
\textbf{GPUs.} We evaluate four GPUs: NVIDIA GeForce RTX~5090 (32\,GB)~\cite{nvidia_rtx5090}, AMD Instinct MI300X (192\,GB)~\cite{amd_mi300x}, NVIDIA H100 (80\,GB)~\cite{nvidia_h100}, and NVIDIA L40 (48\,GB)~\cite{nvidia_l40}.

\textbf{Models and datasets.} We use Qwen3-4B~\cite{yang2025qwen3} target model and its DFlash drafter~\cite{chen2026dflash} with block size 16. Prompts are sampled from GSM8K~\cite{cobbe2021gsm8k}, MATH-500~\cite{lightman2023verify}, MBPP~\cite{austin2021mbpp}, MT-Bench~\cite{zheng2023mtbench}, and HumanEval~\cite{chen2021humaneval}. 



\textbf{Baselines.} We compare against AR and default SD ($k=15$, no adjustments made to default drafting length) served by vLLM. Furthermore, we evaluate against MagicDec \ref{fig:magicdec-comparison}.


\textbf{Calibration.} When the cost model is disabled, we evaluate eight policies across several GPUs: AR decoding with FP16 or FP8 KV cache, and SD with $k\in\{3,7,15\}$, each using FP16 or FP8 KV cache. We consider workloads with $B\in\{8,32,64,128\}$ and $\ell\in\{256,1{,}024,2{,}048\}$. On each GPU, the 12 workloads and three seeds yield $8\times12\times3=288$ calibration runs. We additionally calibrate two dynamic draft-length mappings per workload and seed, adding 72 runs. Thus, an exhaustive calibration performs 360 runs per GPU.

\begin{table*}[t]\centering\scriptsize
\caption{\textbf{Throughput (k tokens/s)} of decoding policies at 15.7\,GiB device budget. \textbf{Bold}: highest mean value per workload. \underline{Underlined}: the policy selected by \sys{}. Within each method (AR, vLLM SD, \sys{}), the darker shading marks fp8 KV in contrast to fp16 KV cache.}\label{tab:sel}
\setlength{\tabcolsep}{2.2pt}
\begin{tabular}{@{}r r A{4}A{10} V{10}V{22} S{8}S{18} rr@{\hspace{8pt}}A{4}A{10} V{10}V{22} S{8}S{18} rr@{}}
\toprule
 & & \multicolumn{8}{c}{MI300X} & \multicolumn{8}{c}{RTX~5090}\\
\cmidrule(lr){3-10}\cmidrule(l){11-18}
 & & \multicolumn{2}{c}{vLLM AR} & \multicolumn{2}{c}{vLLM SD} & \multicolumn{2}{c}{\sys{}} & \multicolumn{2}{c}{Improvement vs.} & \multicolumn{2}{c}{AR} & \multicolumn{2}{c}{vLLM SD} & \multicolumn{2}{c}{\sys{}} & \multicolumn{2}{c}{Improvement vs.}\\
\cmidrule(lr){3-4}\cmidrule(lr){5-6}\cmidrule(lr){7-8}\cmidrule(lr){9-10}\cmidrule(lr){11-12}\cmidrule(lr){13-14}\cmidrule(lr){15-16}\cmidrule(l){17-18}
$\ell$ & $B$ & fp16 & fp8 & fp16 & fp8 & fp16 & fp8 & best AR & best vLLM SD & fp16 & fp8 & fp16 & fp8 & fp16 & fp8 & best AR & best SD\\
\midrule
\multirow{4}{*}{256} & 8 & 1.88 & 1.90 & 4.26 & 4.55 & 4.32 & \underline{\textbf{4.69}} & $2.47\times$ & $1.03\times$ & 1.02 & 1.01 & \textbf{3.91} & 3.89 & \underline{\textbf{3.91}} & 3.84 & $3.83\times$ & $1.00\times$\\
 & 32 & 5.15 & 5.75 & 7.18 & 8.14 & 8.38 & \underline{\textbf{8.88}} & $1.54\times$ & $1.09\times$ & 3.65 & 3.67 & 5.13 & 5.19 & \underline{7.36} & \textbf{7.54} & $2.01\times$ & $1.42\times$\\
 & 64 & 8.21 & 9.34 & 8.45 & 8.93 & 10.88 & \underline{\textbf{11.91}} & $1.28\times$ & $1.33\times$ & 6.37 & 6.65 & 5.44 & 5.53 & \underline{\textbf{9.46}} & 9.31 & $1.42\times$ & $1.71\times$\\
 & 128 & 11.98 & 13.91 & 9.38 & 10.13 & 14.99 & \underline{\textbf{16.24}} & $1.17\times$ & $1.60\times$ & 10.01 & \underline{10.76} & 5.67 & 5.78 & 10.48 & \textbf{10.89} & $1.00\times$ & $1.86\times$\\
\cmidrule(l){1-18}
\multirow{4}{*}{1{,}024} & 8 & 1.81 & 1.86 & 3.58 & 4.18 & 3.58 & \underline{\textbf{4.32}} & $2.32\times$ & $1.03\times$ & 1.03 & 1.03 & 3.73 & \textbf{3.75} & \underline{3.73} & \textbf{3.75} & $3.62\times$ & $0.99\times$\\
 & 32 & 3.92 & 4.84 & 6.09 & 6.89 & 6.85 & \underline{\textbf{7.84}} & $1.62\times$ & $1.14\times$ & 3.25 & 3.31 & 4.89 & 4.94 & \underline{\textbf{7.01}} & 7.00 & $2.12\times$ & $1.42\times$\\
 & 64 & 5.85 & 7.16 & 7.21 & 8.53 & 8.78 & \underline{\textbf{10.15}} & $1.42\times$ & $1.19\times$ & 5.14 & 5.37 & 5.30 & 5.37 & 8.31 & \underline{\textbf{8.53}} & $1.59\times$ & $1.59\times$\\
 & 128 & 8.32 & 10.76 & 8.14 & 10.13 & 9.83 & \underline{\textbf{14.77}} & $1.37\times$ & $1.46\times$ & 7.97 & 9.04 & 5.29 & 5.77 & 8.48 & \underline{\textbf{10.29}} & $1.14\times$ & $1.78\times$\\
\cmidrule(l){1-18}
\multirow{4}{*}{2{,}048} & 8 & 1.75 & 1.81 & 3.42 & \textbf{4.03} & 3.40 & \underline{\textbf{4.03}} & $2.23\times$ & $1.00\times$ & 1.01 & 1.01 & \textbf{3.75} & 3.65 & \underline{\textbf{3.75}} & 3.65 & $3.71\times$ & $1.00\times$\\
 & 32 & 3.14 & 3.93 & 5.50 & 6.47 & 6.05 & \underline{\textbf{7.17}} & $1.82\times$ & $1.11\times$ & 2.67 & 2.77 & 4.91 & 4.81 & 6.12 & \underline{\textbf{6.34}} & $2.29\times$ & $1.29\times$\\
 & 64 & 4.62 & 5.73 & 6.39 & 7.48 & 7.48 & \underline{\textbf{9.19}} & $1.60\times$ & $1.23\times$ & 3.95 & 4.21 & 5.06 & 5.16 & 7.10 & \underline{\textbf{7.65}} & $1.82\times$ & $1.48\times$\\
 & 128 & 6.05 & 8.79 & 6.90 & 8.91 & 8.75 & \underline{\textbf{12.03}} & $1.37\times$ & $1.35\times$ & 6.30 & 7.16 & 5.16 & 5.55 & 7.86 & \underline{\textbf{9.53}} & $1.33\times$ & $1.72\times$\\
\midrule
\multicolumn{2}{c}{Average} & \multicolumn{6}{c}{} & $1.68\times$ & $1.21\times$ & \multicolumn{6}{c}{} & $2.16\times$ & $1.44\times$\\
\bottomrule
\end{tabular}
\end{table*}

\begin{table*}[t]\centering\scriptsize
\caption{\textbf{Speedup} of decoding policies at 15.7\,GiB device budget. \textbf{Bold}: highest mean value per workload. \underline{Underlined}: the policy selected by \sys{}. Within each method (AR, vLLM SD, \sys{}), the darker shading marks fp8 KV in contrast to fp16 KV cache.}\label{tab:speedup}
\setlength{\tabcolsep}{2.2pt}
\begin{tabular}{@{}r r A{4}A{10} V{10}V{22} S{8}S{18} rr@{\hspace{8pt}}A{4}A{10} V{10}V{22} S{8}S{18} rr@{}}
\toprule
 & & \multicolumn{8}{c}{MI300X} & \multicolumn{8}{c}{RTX~5090}\\
\cmidrule(lr){3-10}\cmidrule(l){11-18}
 & & \multicolumn{2}{c}{vLLM AR} & \multicolumn{2}{c}{vLLM SD} & \multicolumn{2}{c}{\sys{}} & \multicolumn{2}{c}{Improvement vs.} & \multicolumn{2}{c}{AR} & \multicolumn{2}{c}{Baseline SD} & \multicolumn{2}{c}{\sys{}} & \multicolumn{2}{c}{Improvement vs.}\\
\cmidrule(lr){3-4}\cmidrule(lr){5-6}\cmidrule(lr){7-8}\cmidrule(lr){9-10}\cmidrule(lr){11-12}\cmidrule(lr){13-14}\cmidrule(lr){15-16}\cmidrule(l){17-18}
$\ell$ & $B$ & fp16 & fp8 & fp16 & fp8 & fp16 & fp8 & best AR & best vLLM SD & fp16 & fp8 & fp16 & fp8 & fp16 & fp8 & best AR & best SD\\
\midrule
\multirow{4}{*}{256} & 8 & 1.00$\times$ & 1.01$\times$ & 2.27$\times$ & 2.42$\times$ & 2.30$\times$ & \underline{\textbf{2.50$\times$}} & $2.46\times$ & $1.03\times$ & 1.00$\times$ & 0.99$\times$ & \textbf{3.84$\times$} & 3.82$\times$ & \underline{\textbf{3.84$\times$}} & 3.77$\times$ & $3.84\times$ & $1.00\times$\\
 & 32 & 1.00$\times$ & 1.12$\times$ & 1.39$\times$ & 1.58$\times$ & 1.62$\times$ & \underline{\textbf{1.72$\times$}} & $1.54\times$ & $1.09\times$ & 1.00$\times$ & 1.01$\times$ & 1.41$\times$ & 1.43$\times$ & \underline{2.02$\times$} & \textbf{2.06$\times$} & $2.00\times$ & $1.41\times$\\
 & 64 & 1.00$\times$ & 1.13$\times$ & 1.03$\times$ & 1.08$\times$ & 1.32$\times$ & \underline{\textbf{1.45$\times$}} & $1.28\times$ & $1.35\times$ & 1.00$\times$ & 1.05$\times$ & 0.85$\times$ & 0.87$\times$ & \underline{\textbf{1.49$\times$}} & 1.46$\times$ & $1.42\times$ & $1.71\times$\\
 & 128 & 1.00$\times$ & 1.16$\times$ & 0.78$\times$ & 0.85$\times$ & 1.25$\times$ & \underline{\textbf{1.36$\times$}} & $1.17\times$ & $1.60\times$ & 1.00$\times$ & \underline{1.08$\times$} & 0.57$\times$ & 0.58$\times$ & 1.05$\times$ & \textbf{1.09$\times$} & $1.00\times$ & $1.87\times$\\
\cmidrule(l){1-18}
\multirow{4}{*}{1{,}024} & 8 & 1.00$\times$ & 1.03$\times$ & 1.99$\times$ & 2.34$\times$ & 1.99$\times$ & \underline{\textbf{2.40$\times$}} & $2.34\times$ & $1.03\times$ & 1.00$\times$ & 1.01$\times$ & 3.63$\times$ & \textbf{3.66$\times$} & \underline{3.63$\times$} & \textbf{3.66$\times$} & $3.59\times$ & $0.99\times$\\
 & 32 & 1.00$\times$ & 1.23$\times$ & 1.57$\times$ & 1.76$\times$ & 1.76$\times$ & \underline{\textbf{2.01$\times$}} & $1.63\times$ & $1.14\times$ & 1.00$\times$ & 1.02$\times$ & 1.53$\times$ & 1.53$\times$ & \underline{\textbf{2.14$\times$}} & 2.13$\times$ & $2.10\times$ & $1.40\times$\\
 & 64 & 1.00$\times$ & 1.22$\times$ & 1.22$\times$ & 1.44$\times$ & 1.50$\times$ & \underline{\textbf{1.72$\times$}} & $1.42\times$ & $1.19\times$ & 1.00$\times$ & 1.05$\times$ & 1.05$\times$ & 1.05$\times$ & \textbf{1.66$\times$} & \underline{\textbf{1.66$\times$}} & $1.58\times$ & $1.59\times$\\
 & 128 & 1.00$\times$ & 1.30$\times$ & 0.98$\times$ & 1.22$\times$ & 1.19$\times$ & \underline{\textbf{1.77$\times$}} & $1.36\times$ & $1.45\times$ & 1.00$\times$ & 1.13$\times$ & 0.67$\times$ & 0.73$\times$ & 1.06$\times$ & \underline{\textbf{1.30$\times$}} & $1.15\times$ & $1.78\times$\\
\cmidrule(l){1-18}
\multirow{4}{*}{2{,}048} & 8 & 1.00$\times$ & 1.06$\times$ & 1.98$\times$ & \textbf{2.32$\times$} & 2.01$\times$ & \underline{\textbf{2.32$\times$}} & $2.19\times$ & $1.00\times$ & 1.00$\times$ & 1.01$\times$ & \textbf{3.77$\times$} & 3.61$\times$ & \underline{\textbf{3.77$\times$}} & 3.61$\times$ & $3.73\times$ & $1.00\times$\\
 & 32 & 1.00$\times$ & 1.23$\times$ & 1.77$\times$ & 2.01$\times$ & 1.93$\times$ & \underline{\textbf{2.26$\times$}} & $1.83\times$ & $1.12\times$ & 1.00$\times$ & 1.01$\times$ & 1.82$\times$ & 1.79$\times$ & 2.26$\times$ & \underline{\textbf{2.30$\times$}} & $2.28\times$ & $1.26\times$\\
 & 64 & 1.00$\times$ & 1.25$\times$ & 1.38$\times$ & 1.60$\times$ & 1.61$\times$ & \underline{\textbf{1.96$\times$}} & $1.57\times$ & $1.22\times$ & 1.00$\times$ & 1.06$\times$ & 1.30$\times$ & 1.28$\times$ & 1.83$\times$ & \underline{\textbf{1.92$\times$}} & $1.81\times$ & $1.47\times$\\
 & 128 & 1.00$\times$ & 1.45$\times$ & 1.13$\times$ & 1.46$\times$ & 1.45$\times$ & \underline{\textbf{1.96$\times$}} & $1.36\times$ & $1.35\times$ & 1.00$\times$ & 1.13$\times$ & 0.83$\times$ & 0.87$\times$ & 1.26$\times$ & \underline{\textbf{1.52$\times$}} & $1.34\times$ & $1.75\times$\\
\midrule
\multicolumn{2}{c}{Average} & \multicolumn{6}{c}{} & $1.68\times$ & $1.21\times$ & \multicolumn{6}{c}{} & $2.15\times$ & $1.44\times$\\
\bottomrule
\end{tabular}
\end{table*}

\section{Results}\label{sec:eval}
In this section, we evaluate both policy selection modes of \sys{}: full calibration and learning-based policy selection. 

In full calibration mode, \sys{} uses its analytical cost model to refine the policy search space and identify feasible candidate configurations, then selects the best-performing policy through GPU-specific calibration. 

The learning-based mode trains a Jev-based predictor to select optimal configurations from calibration data. As a result, the accuracy and portability of the cost model directly affect the quality of full calibration policy selection and the training dataset of the learned predictor. 

We first validate the cost model across GPUs by comparing its predictions against measurement-based reference points obtained from actual vLLM executions. We then evaluate the end-to-end performance of both policy selection modes.

\subsection{Cost model validation across GPUs}\label{sec:costval}
Table~\ref{tab:cost} validates the proposed cost model, including memory, concurrency, and throughput. We compare the predicted cost by our model against the measured cost when running vLLM for AR and SD with two KV cache precisions. Eq.~\eqref{eq:mem} predicts total engine memory within $0.1\%$ on the MI300X and $0.2\%$ on the RTX~5090. Equation~\eqref{eq:cap} predicts maximum concurrency within $0.5\%$ and $0.6\%$, respectively. Across the tested budgets, the cost model correctly rejects 7 of 20 configuration--budget pairs on the MI300X and 3 of 20 on the RTX~5090. These rejected pairs occur at allocations below the 15.7\,GiB budget used for the main evaluation. In other words, the cost model correctly accepts all remaining pairs.

The model-dependent per-token KV-cache cost $(\kappa)$ and drafter weight footprint match measured allocations within $1\%$, without any device-specific adjustments. Only the constant runtime overheads ($\rho_g$, $C_{\text{act}}$, and $C_{\text{graph}}$) require a one-time startup measurement on each GPU. These results validate the cost model on MI300X and RTX~5090 after device-specific initialization.


\begin{figure}[htbp]
\centering
\includegraphics[width=0.9\columnwidth]{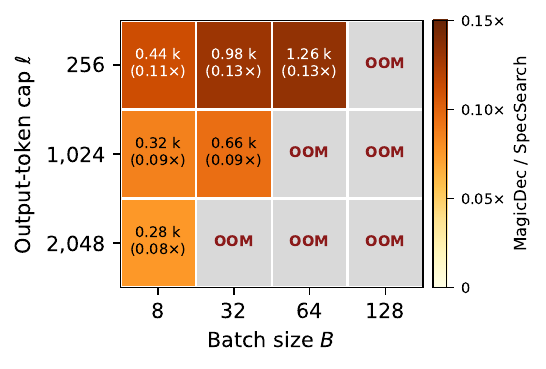}
\caption{Throughput (in k token/s) comparison of MagicDec vs. \sys{} for Qwen-3-4B. Each cell reports MagicDec throughput on RTX5090, while its ratio to \sys{} is given in parentheses. Gray cells denote out-of-memory (OOM) failures.}
\label{fig:magicdec-comparison}
\end{figure}

\subsection{Performance of \sys{} with full calibration}\label{sec:sel}

\textbf{Throughput and speedup results.}
Table~\ref{tab:sel} and~\ref{tab:speedup} report throughput (in k tokens/s) and wall-clock speedup measured by the calibration runs, respectively. The last two columns in each GPU block report the throughput improvements achieved by the policy \sys{} selects over the fastest AR and vLLM SD configurations, respectively. Each GPU has available 15.7 GiB device budget for LLM inference. Similarly, Table~\ref{tab:sel-appendix} and~\ref{tab:speedup-appendix} report findings on NVIDIA H100 and L40 GPUs at a 15.7 GiB budget. 

The performance gains vary with concurrency, output length, and GPU. At $B=8$, vLLM SD is already competitive, leaving little room for further improvement (Table~\ref{tab:sel}). At higher concurrency, \sys{} more consistently outperforms the vLLM SD baseline, particularly for longer outputs. For example, at $B=128$ and $\ell=2{,}048$, \sys{} reaches 12.03\,k tokens/s on the MI300X and 9.53\,k tokens/s on the RTX~5090, versus 8.91 and 5.55\,k tokens/s, respectively, for vLLM SD with fp8 KV. The same setup delivers wall-clock speedups of $1.96\times$ and $1.52\times$ over AR with fp16 KV. 

Averaged across workloads, \sys{} achieves throughput improvements of $1.68\times$, $2.16\times$, $2.33\times$, and $1.85\times$ over the best AR configuration on the MI300X, RTX~5090, H100, and L40, respectively. The corresponding speedup values over vLLM SD are $1.21\times$, $1.44\times$, $1.28\times$, and $1.51\times$. These results demonstrate \sys{}'s ability to identify the best deployment configuration depending on both the workload and the GPU.

Figure~\ref{fig:magicdec-comparison} compares the end-to-end throughput of MagicDec and \sys{} on the same RTX~5090, using Qwen3-4B as the target model and an identical prompt set. The two systems employ different serving optimizations: MagicDec uses sparse-KV self-speculation to compress the KV cache and reduce memory consumption for long-context workloads, with BF16 KV cache, static batching, and no request refill in our evaluation. In contrast, \sys{} uses the vLLM backend, which supports paged KV caching and continuous batching, and jointly selects the KV-cache precision and speculation policy. These differences in serving mechanisms also affect the set of workloads each system can execute: MagicDec cannot run six decoding configurations due to memory constraints. Among the workloads supported by both systems, MagicDec achieves 0.28--1.26\,k tokens/s, corresponding to $0.08$--$0.13\times$ the throughput of \sys{}. The results thus compare the end-to-end performance of the respective system designs under a common hardware, model, and prompt configuration, rather than isolating the contribution of individual optimizations.

\begin{table}[!htbp]\centering\scriptsize
\caption{Policy selected by \sys{} in Table \ref{tab:sel} and \ref{tab:speedup}: AR vs. SD decoding (with draft length $k$) and KV precision. $k^{\dagger}$: dynamic draft-length mapping for SD (Table~\ref{tab:maps}).}\label{tab:selpol}
\setlength{\tabcolsep}{4pt}
\begin{tabular}{@{}l r cccc@{}}
\toprule
 & $\ell$ & $B{=}8$ & $B{=}32$ & $B{=}64$ & $B{=}128$\\
\midrule
\multirow{3}{*}{MI300X} & 256 & $k{=}7$, fp8 & $k{=}3$, fp8 & $k{=}3$, fp8 & $k{=}3$, fp8\\
 & 1{,}024 & $k{=}7$, fp8 & $k{=}3$, fp8 & $k{=}3$, fp8 & $k{=}3$, fp8\\
 & 2{,}048 & $k^{\dagger}$, fp8 & $k{=}7$, fp8 & $k{=}7$, fp8 & $k{=}3$, fp8\\
\midrule
\multirow{3}{*}{RTX~5090} & 256 & $k{=}15$, fp16 & $k{=}3$, fp16 & $k{=}3$, fp16 & AR, fp8\\
 & 1{,}024 & $k{=}15$, fp16 & $k^{\dagger}$, fp16 & $k{=}3$, fp8 & $k{=}3$, fp8\\
 & 2{,}048 & $k{=}15$, fp16 & $k^{\dagger}$, fp8 & $k^{\dagger}$, fp8 & $k{=}3$, fp8\\
\midrule
\multirow{3}{*}{H100} & 256 & $k^{\dagger}$, fp8 & $k^{\dagger}$, fp8 & $k{=}3$, fp8 & $k{=}3$, fp16\\
 & 1{,}024 & $k^{\dagger}$, fp16 & $k^{\dagger}$, fp8 & $k{=}3$, fp8 & $k{=}3$, fp8\\
 & 2{,}048 & $k{=}15$, fp16 & $k^{\dagger}$, fp8 & $k{=}7$, fp16 & $k{=}3$, fp8\\
\midrule
\multirow{3}{*}{L40} & 256 & $k^{\dagger}$, fp8 & $k{=}3$, fp8 & $k{=}3$, fp8 & AR, fp8\\
 & 1{,}024 & $k^{\dagger}$, fp8 & $k{=}3$, fp8 & $k{=}3$, fp8 & $k{=}3$, fp8\\
 & 2{,}048 & $k{=}15$, fp8 & $k{=}3$, fp8 & $k{=}3$, fp8 & $k{=}3$, fp8\\
\bottomrule
\end{tabular}
\end{table}

\textbf{Selected policies.} \sys{} calibrates SD across different draft lengths to construct dynamic draft-length mappings. Table~\ref{tab:selpol} details the policies selected by \sys{} in Tables~\ref{tab:sel} and~\ref{tab:speedup}, including their corresponding draft lengths. Tables~\ref{tab:fixedk-throughput} and~\ref{tab:fixedk-speedup} provide the individual calibration results used to derive draft length mappings. On the MI300X, \sys{} selects speculation with fp8 KV at every workload, using a dynamic draft-length mapping at one workload and a static draft length at eleven workloads. On the RTX~5090, it selects fp16 KV at six workloads and fp8 KV at six, including the one workload at which AR decoding is selected. Eight RTX~5090 workloads use a static draft length, three use the dynamic mapping, and one uses AR decoding. Table~\ref{tab:maps} lists the dynamic draft length mappings constructed by \sys{}. On H100, five workloads use dynamic mappings and seven use static draft lengths, with fp16 KV selected at four workloads and fp8 at eight. On L40, two workloads use dynamic mappings, nine use static draft lengths, while one uses AR and all select fp8 KV. 
\begin{table}[!htbp]\centering\scriptsize
\caption{Dynamic draft-length mappings of \sys{}. Each entry reports the selected draft length $k$ for fp16/fp8 KV-cache precision, respectively, at the indicated effective-batch range.}
\label{tab:maps}
\setlength{\tabcolsep}{4pt}
\begin{tabular}{@{}l l cccc@{}}
\toprule
GPU & $\ell$ & 1--20 & 21--48 & 49--96 & 97--128\\
\midrule
\multirow{3}{*}{MI300X} & 256 & 7/7 & 7/3 & 3/3 & 3/3\\
 & 1{,}024 & 15/15 & 7/3 & 7/3 & 3/3\\
 & 2{,}048 & 15/15 & 7/7 & 7/3 & 7/3\\
\midrule
\multirow{3}{*}{RTX~5090}
 & 256     & 15/15 & 3/3 & 3/3 & 3/3\\
 & 1{,}024 & 15/15 & 3/3 & 3/3 & 3/3\\
 & 2{,}048 & 15/15 & 7/7 & 3/3 & 3/3\\
\midrule
\multirow{3}{*}{H100} & 256 & 15/15 & 7/7 & 3/3 & 3/3\\
 & 1{,}024 & 15/15 & 7/7 & 3/3 & 3/3\\
 & 2{,}048 & 15/15 & 7/7 & 7/7 & 3/3\\
\midrule
\multirow{3}{*}{L40} & 256 & 15/15 & 3/3 & 3/3 & 3/3\\
 & 1{,}024 & 15/15 & 3/3 & 3/3 & 3/3\\
 & 2{,}048 & 15/15 & 3/3 & 3/3 & 3/3\\
\bottomrule
\end{tabular}
\end{table}
\begin{table}[hbtp]
\centering
\scriptsize
\caption{Quantization explored under an 8\,GiB memory budget on the RTX 5090. Shaded rows indicate policies that exceed the device memory budget. Target and draft denote the target and draft model precisions, respectively. $\ell=4{,}096$.}
\label{tab:b8}
\setlength{\tabcolsep}{3pt}
\resizebox{\columnwidth}{!}{%
\begin{tabular}{@{}l c c c r r r r r r@{}}
\toprule
policy & target & draft & KV cache & weights (GiB) & KV tokens & max conc. & ktok/s & wall (s) & speedup\\
\midrule
\rowcolor{gray!15}
AR & fp16 & -- & fp16 & 7.56 & \multicolumn{5}{l}{}\\
\rowcolor{gray!15}
AR & fp16 & -- & fp8 & 7.59 & \multicolumn{5}{l}{}\\
AR & fp8 & -- & fp16 & 4.24 & 19{,}712 & $4.81\times$ & 8.617 & 5.942 & $0.63\times$\\
AR & fp8 & -- & fp8 & 4.24 & 39{,}568 & $9.66\times$ & \textbf{13.661} & \textbf{3.748} & $\mathbf{1.00\times}$\\
\midrule
\rowcolor{gray!15}
SD & fp16 & fp16 & fp16 & 8.61 & \multicolumn{5}{l}{}\\
\rowcolor{gray!15}
SD & fp16 & fp16 & fp8 & 8.65 & \multicolumn{5}{l}{}\\
\rowcolor{gray!15}
SD & fp8 & fp16 & fp16 & 5.36 & \multicolumn{5}{l}{}\\
SD & fp8 & fp16 & fp8 & 5.36 & 5{,}376 & $1.31\times$ & 5.701 & 8.981 & $0.42\times$\\
\rowcolor{gray!15}
SD & fp16 & fp8 & fp8 & 8.1 & \multicolumn{5}{l}{}\\
SD & fp8 & fp8 & fp8 & 4.83 & 11{,}808 & $2.88\times$ & 5.898 & 8.681 & $0.43\times$\\
\rowcolor{gray!15}
SD & fp16 & int4 & fp8 & 7.85 & \multicolumn{5}{l}{}\\
SD & fp8 & int4 & fp8 & 4.55 & 6{,}960 & $1.70\times$ & 6.093 & 8.403 & $0.45\times$\\
\bottomrule
\end{tabular}
}
\end{table}
\textbf{Target model quantization.} At a smaller device budget such as 8 GB, serving SD may require quantizing the target model. Table~\ref{tab:b8} shares the policies with a quantized target and drafter that calibration explores, demonstrating feasible configurations under this budget on an RTX~5090. The SD policies that fit 8 GB are slower than AR decoding on this GPU. Table~\ref{tab:tq} reports the throughput of configurations with an fp8 target explored by \sys{} on the MI300X, using a 15.7 GiB budget. Within this setting, \sys{} averages $2.17\times$ the throughput of AR and $1.03\times$ that of vLLM SD. Relative to the corresponding fp16-target results in Table~\ref{tab:sel}, target quantization reduces the throughput of AR, vLLM SD, and \sys{} across all workloads on the MI300X GPU. This is because quantizing the target model perturbs the reference distribution, introducing loss to verification and harming the overall acceptance.

\textbf{Drafter model quantization.} Quantized drafter models leave the output distribution of SD unchanged, maintaining losslessness. Since drafter only determines the draft token quality, a quantized drafter only impacts the draft token acceptance length. Table~\ref{tab:asym} reports this effect, measured on the RTX~5090 at $\ell{=}2{,}048$ with fp16 target and fp8 KV cache. Quantized fp8 and int4 drafters reduce drafter weight memory by $50\%$ and $74\%$, while the corresponding acceptance retention stays within $1\%$ of the fp16 drafter. Thus, neither drafter precision changes serving throughput measurably.

\begin{table}[!htbp]\centering\footnotesize
\caption{Drafter quantization on the RTX~5090. $\bar\tau$ is the mean acceptance length of draft tokens across datasets. Retention is the rate at which mean acceptance and throughput are preserved relative to fp16.}
\label{tab:asym}
\setlength{\tabcolsep}{3.5pt}
\begin{tabular}{@{}r l r rr rr@{}}
\toprule
batch size & drafter & weights & \multicolumn{2}{c}{acceptance} & \multicolumn{2}{c}{throughput}\\
\cmidrule(lr){4-5}\cmidrule(lr){6-7}
 & & (GiB) & $\bar\tau$ & retention & k tok/s & retention\\
\midrule
\multirow{3}{*}{32}
& fp16 & 1.01 & 9.87 & $1.0$ & 8.98 & $1.0\times$\\
& fp8 & 0.51 & 9.90 & $1.0$ & 9.16 & $1.0\times$\\
& int4 & 0.26 & 9.81 & $0.9$ & 9.00 & $1.0\times$\\
\midrule
\multirow{3}{*}{64}
& fp16 & 1.01 & 9.81 & $1.0$ & 8.63 & $1.0\times$\\
& fp8 & 0.51 & 10.03 & $1.0$ & 9.43 & $1.0\times$\\
& int4 & 0.26 & 9.76 & $0.9$ & 8.97 & $1.0\times$\\
\bottomrule
\end{tabular}
\end{table}
\subsection{Performance of Learning-Based Policy Predictor}\label{sec:learning-eval}
We evaluate zero-shot policy selection using a four-fold cross-GPU evaluation. In each fold, one of the four GPUs is held out for evaluation, while measurements from the remaining three GPUs are used to train the predictor. We repeat this process four times, using each GPU as the held-out GPU exactly once. Each GPU contributes 12 workloads, resulting in 48 workload-hardware pairs across the four folds. We assess each policy predicted by our Jev-based selector by measuring its throughput on the held-out GPU. These measurements are strictly reserved for evaluation and are excluded from predictor training and model selection. We compare the performance of configurations selected by the Jev-based predictor against the performance of policy search predicted in full calibration mode. Unlike the learned predictor, this \sys{} mode has access to calibration measurements from the evaluation GPU.

\textbf{Selected policy performance.} Table~\ref{tab:learning-eval} evaluates how effectively the trained Jev-based predictor selects decoding configurations on unseen GPUs, comparing it against the baseline vLLM decoding. The Jev-based predictor achieves average throughput ratios of $2.02\times$ over AR and $1.30\times$ over vLLM SD. Selected policies are shared in Appendix \ref{app:transfer-maps}.

\textbf{Comparison with full calibration.} Table~\ref{tab:learning-eval} shares the throughput gap that measures the relative performance loss of the inferred policy compared with the best-performing configuration returned by \sys{} in full calibration mode. A smaller gap indicates more efficient policy selection. Relative to the best measured static configuration, its mean throughput loss is $0.75\%$ on RTX~5090, $1.80\%$ on MI300X, $1.52\%$ on L40, and $5.95\%$ on H100. Averaged equally across the four held-out GPUs, the fine-tuned Jev-based predictor retains $97.36\%$ of the matched full calibration \sys{} throughput. Jev-based fine-tuning maintains an average throughput gap of $2.50\%$ to the optimal policies discovered during calibration. These results demonstrate effective policy selection without deployment-GPU profiling.

\textbf{Predictions across varying concurrency.} We study how workload concurrency influences the predictor's choices. The largest benefit of Jev-based prediction occurs at $B{=}8$, where the mean throughput gap to the best configuration's throughput decreases from $10.76\%$ to $1.82\%$. The Jev-based model selects $k{=}15$ for all 12 workloads at this concurrency, whereas the frozen Qwen-3-4B decision model selects $k{=}3$ for all four 256-token workloads. At $B{=}64$, both models sets $k{=}3$ with fp8 KV cache precision. At $B{=}128$, the Jev-based model selects AR with fp8 KV for the 256-token RTX~5090 workload and $k{=}3$ for the other 11 workloads. These choices reflect the load-dependent behavior observed in Section~\ref{sec:motiv}: longer drafts remain beneficial at low concurrency, whereas shorter drafts or AR remain faster as concurrency increases.

\textbf{Prediction errors.} The Jev-based predictor selects a configuration within $5\%$ of the best configuration's throughput for 38 of the 48 workloads. Most of the larger prediction errors occur on H100 at $B{=}32$: the predictor selects $k{=}3$ with fp8 KV for all output lengths from 256 to 2048 tokens. For these workloads, the best-performing configurations use $k{=}7$ on H100. The resulting throughput gaps range from $13.73\%$ to $21.64\%$. These results reveal a limitation at moderate concurrency on H100, where the predictor favors shorter drafts than the measured throughput optimum.

\begin{table}[!t]
\centering
\footnotesize
\caption{Learned policy selection via Jev-based predictor. First row reports its throughput gap compared to the best policy returned by full calibration (F.C.) \sys{}. Last two rows are speedup over baseline vLLM runs.}
\label{tab:learning-eval}
\setlength{\tabcolsep}{3pt}
\begin{adjustbox}{max width=\columnwidth}
\begin{tabular}{@{}l rrrr r@{}}
\toprule
& \multicolumn{4}{c}{GPU} & \\
\cmidrule(lr){2-5}
& RTX~5090 & MI300X & H100 & L40 & Mean\\
\midrule
Jev-based vs. F.C. & \textbf{0.75\%} & 1.80\% & 5.95\% & 1.52\% & 2.50\%\\
\midrule
Jev-based vs. vLLM AR & $2.24\times$ & $1.68\times$ & \textbf{2.27$\times$} & $1.90\times$ & $2.02\times$\\
Jev-based vs. vLLM SD & $1.40\times$ & $1.14\times$ & $1.19\times$ & \textbf{1.47$\times$} & $1.30\times$\\
\bottomrule
\end{tabular}
\end{adjustbox}
\end{table}

\begin{figure}[h!]\centering
\includegraphics[width=0.9\columnwidth]{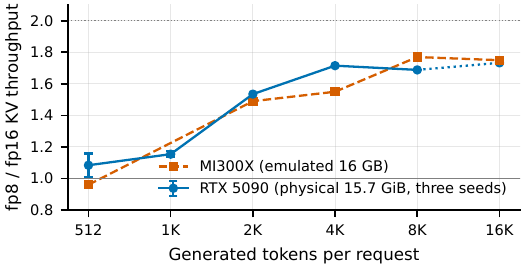}
\caption{AR decoding: fp8-to-fp16 KV throughput ratio versus generated tokens per request, with concurrency $48$.}\label{fig:fp8len}
\end{figure}

\subsection{Ablations}\label{sec:ablations}
\textbf{Memory cost of speculation.}
Fig.~\ref{fig:port}(c,d) and Table~\ref{tab:fair} quantify the memory cost of speculation described in Observation~3. At equal KV precision, vLLM SD retains $0.73\times$ the AR KV cache capacity on the RTX~5090 under a 15.7\,GiB budget and $0.68\times$ on the MI300X under a 16\,GB budget. Despite its smaller KV cache capacity, vLLM SD delivers $3.5$--$3.9\times$ the AR throughput at concurrency 1--4. This advantage disappears by concurrency 64, where the throughput ratio falls to $0.93$--$0.97\times$.

\textbf{Generation length sensitivity.} An fp8 KV cache doubles the KV cache capacity of both decoding methods. For short outputs, however, it changes throughput by at most $4\%$. The additional capacity becomes useful as outputs grow longer. For AR decoding on the RTX~5090, the fp8-to-fp16 throughput ratio rises from $1.08\times$ at 512 generated tokens to $1.73\times$ at 16,384 tokens (Fig.~\ref{fig:fp8len}). The MI300X exhibits the same trend.

\begin{table}[t]\centering\footnotesize
\caption{Relationship between KV cache capacity, in tokens, and maximum concurrency. The workload has $\ell = 4,096$ across runs.}\label{tab:fair}
\setlength{\tabcolsep}{3pt}
\begin{tabular}{lrrrr}
\toprule
& \multicolumn{2}{c}{RTX~5090} & \multicolumn{2}{c}{MI300X}\\
\cmidrule(lr){2-3} \cmidrule(lr){4-5}
policy & \shortstack{capacity\\(tokens)} & \shortstack{max conc.\\(queries)} & \shortstack{capacity\\(tokens)} & \shortstack{max conc.\\(queries)}\\
\midrule
AR, fp16 KV & 54{,}800 & $13.4\times$ & 49{,}632 & $12.1\times$\\
AR, fp8 KV & 109{,}168 & $26.7\times$ & 99{,}264 & $24.2\times$\\
vLLM SD, $k{=}15$, fp16 KV & 40{,}000 & $9.8\times$ & 33{,}808 & $8.3\times$\\
vLLM SD, $k{=}15$, fp8 KV & 79{,}600 & $19.4\times$ & 67{,}632 & $16.5\times$\\
\midrule
Concurrency retention rate & \multicolumn{2}{c}{$0.73$} & \multicolumn{2}{c}{$0.68$}\\
\bottomrule
\end{tabular}
\end{table}

\begin{table}[t]\centering\scriptsize
\caption{Throughput (k tokens/s) with target model quantization (fp8) on the MI300X. \sys{} uses a dynamic draft-length mapping with the fp8 target. The improvement columns report \sys{} throughput relative to the vLLM baseline. \textbf{Bold}: highest mean throughput per workload.}\label{tab:tq}
\setlength{\tabcolsep}{3pt}
\begin{tabular}{@{}r r rrr rr@{}}
\toprule
 & & & & & \multicolumn{2}{c}{Improvement vs.}\\
$\ell$ & $B$ & AR & vLLM SD & \sys{} & best AR & vLLM SD\\
\midrule
\multirow{4}{*}{256} & 8 & 0.86 & \textbf{3.04} & 2.69 & $3.14\times$ & $0.89\times$\\
 & 32 & 2.97 & \textbf{6.42} & 6.27 & $2.11\times$ & $0.98\times$\\
 & 64 & 4.99 & 7.77 & \textbf{8.74} & $1.75\times$ & $1.12\times$\\
 & 128 & 7.48 & 8.49 & \textbf{9.98} & $1.33\times$ & $1.18\times$\\
\cmidrule(l){1-7}
\multirow{4}{*}{1{,}024} & 8 & 0.83 & \textbf{2.88} & 2.87 & $3.47\times$ & $1.00\times$\\
 & 32 & 2.59 & \textbf{5.44} & 5.18 & $2.00\times$ & $0.95\times$\\
 & 64 & 3.96 & 6.26 & \textbf{6.49} & $1.64\times$ & $1.04\times$\\
 & 128 & 5.25 & 6.71 & \textbf{7.56} & $1.44\times$ & $1.13\times$\\
\cmidrule(l){1-7}
\multirow{4}{*}{2{,}048} & 8 & 0.83 & \textbf{2.81} & 2.69 & $3.24\times$ & $0.96\times$\\
 & 32 & 2.24 & 5.01 & \textbf{5.48} & $2.45\times$ & $1.09\times$\\
 & 64 & 3.36 & \textbf{5.89} & 5.74 & $1.71\times$ & $0.97\times$\\
 & 128 & 3.84 & 6.31 & \textbf{6.87} & $1.79\times$ & $1.09\times$\\
\midrule
\multicolumn{5}{c}{Average} & $2.17\times$ & $1.03\times$\\
\bottomrule
\end{tabular}
\end{table}

\section{Limitations}\label{sec:limits}
This study focuses on a single target model and a single diffusion-based drafter. Although the proposed cost model is portable across model architectures, extending it to alternative SD methods, such as EAGLE~\cite{li2024eagle,li2025eagle3} and Medusa~\cite{cai2024medusa}, requires adapting the terms that characterize drafter-specific costs. The empirical results remain limited to the chosen DFlash drafter.

\sys{} supports dynamic draft-length policies by deriving them from calibration runs over static draft lengths. However, the learned Jev-based predictor currently selects only static configurations. Incorporating calibration measurements for dynamic draft-length policies into the training dataset could extend the predictor to support dynamic policy selection. We leave this as future work.

The learned Jev-based predictor also exhibits performance gaps on certain workload--GPU combinations, most notably at moderate concurrency on H100, where it favors shorter drafts than the measured throughput optimum. Future work could improve prediction accuracy by expanding the training data across a broader range of GPU architectures and workloads, and by incorporating more informative hardware and workload features into the decision prompts.

\section{Conclusion}
Speculative decoding has no universally optimal configuration, as its performance depends on workload characteristics and GPU hardware. For example, on the RTX~5090, its speedup over AR drops from $3.50\times$ at low concurrency to $0.76\times$ at high concurrency when generation length shortens. To address this challenge, \sys{} combines an analytical cost model that narrows the deployment search space with two complementary policy-selection approaches. full calibration \sys{} evaluates candidate configurations on the deployment GPU, jointly optimizing KV-cache and model-weight precision, draft length, and whether to enable speculation. It achieves average throughput improvements of $2.09\times$ over AR and $1.36\times$ over vLLM SD across four GPUs.

Using the calibration dataset, \sys{} fine-tunes a Jev-based optimal policy predictor. Across four held-out GPUs, it maintains a minimal throughput gap of $2.50\%$ to the best-performing configurations returned by full calibration \sys{}. The learned policy predictor achieves $2.02\times$ the throughput of AR and $1.30\times$ that of vLLM SD, retaining $97.36\%$ of the throughput achieved by full calibration \sys{}. 

Together, these approaches enable hardware-aware policy selection across diverse workloads and deployment conditions, improving inference-serving efficiency on resource-constrained GPUs.
\clearpage

\raggedbottom
\bibliographystyle{IEEEtran}
\bibliography{references}

@misc{leviathan2023fast,
  title         = {{Fast Inference from Transformers via Speculative Decoding}},
  author        = {Yaniv Leviathan and Matan Kalman and Yossi Matias},
  year          = {2022},
  eprint        = {2211.17192},
  archivePrefix = {arXiv},
  howpublished  = {arXiv:2211.17192}
}

@misc{chen2023accelerating,
  title         = {{Accelerating Large Language Model Decoding with Speculative Sampling}},
  author        = {Charlie Chen and Sebastian Borgeaud and Geoffrey Irving and Jean-Baptiste Lespiau and Laurent Sifre and John Jumper},
  year          = {2023},
  eprint        = {2302.01318},
  archivePrefix = {arXiv},
  howpublished  = {arXiv:2302.01318}
}

@misc{cai2024medusa,
  title         = {{Medusa: Simple LLM Inference Acceleration Framework with Multiple Decoding Heads}},
  author        = {Tianle Cai and Yuhong Li and Zhengyang Geng and Hongwu Peng and Jason D. Lee and Deming Chen and Tri Dao},
  year          = {2024},
  eprint        = {2401.10774},
  archivePrefix = {arXiv},
  howpublished  = {arXiv:2401.10774}
}

@misc{li2024eagle,
  title         = {{EAGLE: Speculative Sampling Requires Rethinking Feature Uncertainty}},
  author        = {Yuhui Li and Fangyun Wei and Chao Zhang and Hongyang Zhang},
  year          = {2024},
  eprint        = {2401.15077},
  archivePrefix = {arXiv},
  howpublished  = {arXiv:2401.15077}
}

@misc{li2025eagle3,
  title         = {{EAGLE-3: Scaling up Inference Acceleration of Large Language Models via Training-Time Test}},
  author        = {Yuhui Li and Fangyun Wei and Chao Zhang and Hongyang Zhang},
  year          = {2025},
  eprint        = {2503.01840},
  archivePrefix = {arXiv},
  howpublished  = {arXiv:2503.01840}
}

@misc{chen2026dflash,
  title         = {{DFlash: Block Diffusion for Flash Speculative Decoding}},
  author        = {Jian Chen and Yesheng Liang and Zhijian Liu},
  year          = {2026},
  eprint        = {2602.06036},
  archivePrefix = {arXiv},
  howpublished  = {arXiv:2602.06036}
}

@misc{yildirim2025specmemo,
  title         = {{SpecMemo: Speculative Decoding is in Your Pocket}},
  author        = {Selin Yildirim and Deming Chen},
  year          = {2025},
  eprint        = {2506.01986},
  archivePrefix = {arXiv},
  howpublished  = {arXiv:2506.01986}
}

@misc{liu2024turbospec,
  title         = {{TurboSpec: Closed-loop Speculation Control System for Optimizing LLM Serving Goodput}},
  author        = {Xiaoxuan Liu and Jongseok Park and Langxiang Hu and Woosuk Kwon and Zhuohan Li and Chen Zhang and Kuntai Du and Xiangxi Mo and Kaichao You and Alvin Cheung and Zhijie Deng and Ion Stoica and Hao Zhang},
  year          = {2024},
  eprint        = {2406.14066},
  archivePrefix = {arXiv},
  howpublished  = {arXiv:2406.14066}
}

@inproceedings{sadhukhan2024magicdec,
  title     = {{MagicDec: Breaking the Latency-Throughput Tradeoff for Long Context Generation with Speculative Decoding}},
  author    = {Ranajoy Sadhukhan and Jian Chen and Zhuoming Chen and Vashisth Tiwari and Ruihang Lai and Jinyuan Shi and Ian En-Hsu Yen and Avner May and Tianqi Chen and Beidi Chen},
  booktitle = {The Thirteenth International Conference on Learning Representations},
  year      = {2025},
  url       = {https://proceedings.iclr.cc/paper_files/paper/2025/hash/13f972adf12bdf886583d48cd528002f-Abstract-Conference.html}
}

@inproceedings{chen2024sequoia,
  title     = {{Sequoia: Scalable and Robust Speculative Decoding}},
  author    = {Zhuoming Chen and Avner May and Ruslan Svirschevski and Yuhsun Huang and Max Ryabinin and Zhihao Jia and Beidi Chen},
  booktitle = {Advances in Neural Information Processing Systems},
  volume    = {37},
  pages     = {129531--129563},
  year      = {2024},
  doi       = {10.52202/079017-4116}
}

@inproceedings{li2026adaserve,
  title     = {{AdaServe: Accelerating Multi-SLO LLM Serving with SLO-Customized Speculative Decoding}},
  author    = {Zikun Li and Zhuofu Chen and Remi Delacourt and Gabriele Oliaro and Zeyu Wang and Qinghan Chen and Shuhuai Lin and April Yang and Zhihao Zhang and Zhuoming Chen and Yi-Hsiang Lai and Xinhao Cheng and Xupeng Miao and Zhihao Jia},
  booktitle = {Proceedings of the 21st European Conference on Computer Systems},
  year      = {2026},
  doi       = {10.1145/3767295.3769315}
}

@misc{liu2024kivi,
  title         = {{KIVI: A Tuning-Free Asymmetric 2bit Quantization for KV Cache}},
  author        = {Zirui Liu and Jiayi Yuan and Hongye Jin and Shaochen Zhong and Zhaozhuo Xu and Vladimir Braverman and Beidi Chen and Xia Hu},
  year          = {2024},
  eprint        = {2402.02750},
  archivePrefix = {arXiv},
  howpublished  = {arXiv:2402.02750}
}

@misc{hooper2024kvquant,
  title         = {{KVQuant: Towards 10 Million Context Length LLM Inference with KV Cache Quantization}},
  author        = {Coleman Hooper and Sehoon Kim and Hiva Mohammadzadeh and Michael W. Mahoney and Yakun Sophia Shao and Kurt Keutzer and Amir Gholami},
  year          = {2024},
  eprint        = {2401.18079},
  archivePrefix = {arXiv},
  howpublished  = {arXiv:2401.18079}
}

@misc{lin2024awq,
  title         = {{AWQ: Activation-aware Weight Quantization for LLM Compression and Acceleration}},
  author        = {Ji Lin and Jiaming Tang and Haotian Tang and Shang Yang and Wei-Ming Chen and Wei-Chen Wang and Guangxuan Xiao and Xingyu Dang and Chuang Gan and Song Han},
  year          = {2023},
  eprint        = {2306.00978},
  archivePrefix = {arXiv},
  howpublished  = {arXiv:2306.00978}
}

@misc{frantar2023gptq,
  title         = {{GPTQ: Accurate Post-Training Quantization for Generative Pre-trained Transformers}},
  author        = {Elias Frantar and Saleh Ashkboos and Torsten Hoefler and Dan Alistarh},
  year          = {2022},
  eprint        = {2210.17323},
  archivePrefix = {arXiv},
  howpublished  = {arXiv:2210.17323}
}

@misc{yang2025qwen3,
  title         = {{Qwen3 Technical Report}},
  author        = {An Yang and others},
  year          = {2025},
  eprint        = {2505.09388},
  archivePrefix = {arXiv},
  howpublished  = {arXiv:2505.09388}
}

@misc{cobbe2021gsm8k,
  title         = {{Training Verifiers to Solve Math Word Problems}},
  author        = {Karl Cobbe and Vineet Kosaraju and Mohammad Bavarian and Mark Chen and Heewoo Jun and Lukasz Kaiser and Matthias Plappert and Jerry Tworek and Jacob Hilton and Reiichiro Nakano and Christopher Hesse and John Schulman},
  year          = {2021},
  eprint        = {2110.14168},
  archivePrefix = {arXiv},
  howpublished  = {arXiv:2110.14168}
}

@misc{lightman2023verify,
  title         = {{Let's Verify Step by Step}},
  author        = {Hunter Lightman and others},
  year          = {2023},
  howpublished  = {arXiv:2305.20050}
}

@misc{austin2021mbpp,
  title         = {{Program Synthesis with Large Language Models}},
  author        = {Jacob Austin and others},
  year          = {2021},
  howpublished  = {arXiv:2108.07732}
}

@misc{zheng2023mtbench,
  title         = {{Judging LLM-as-a-Judge with MT-Bench and Chatbot Arena}},
  author        = {Lianmin Zheng and others},
  year          = {2023},
  howpublished  = {arXiv:2306.05685}
}

@misc{chen2021humaneval,
  title         = {{Evaluating Large Language Models Trained on Code}},
  author        = {Mark Chen and Jerry Tworek and Heewoo Jun and Qiming Yuan and Henrique Ponde de Oliveira Pinto and Jared Kaplan and others},
  year          = {2021},
  howpublished  = {arXiv:2107.03374}
}

@inproceedings{kwon2023vllm, title={Efficient Memory Management for Large Language Model Serving with PagedAttention}, url={http://dx.doi.org/10.1145/3600006.3613165}, DOI={10.1145/3600006.3613165}, booktitle={Proceedings of the 29th Symposium on Operating Systems Principles}, publisher={ACM}, author={Kwon, Woosuk and Li, Zhuohan and Zhuang, Siyuan and Sheng, Ying and Zheng, Lianmin and Yu, Cody Hao and Gonzalez, Joseph and Zhang, Hao and Stoica, Ion}, year={2023}, month=Oct, pages={611--626} }

@article{williams2009roofline, title={Roofline: an insightful visual performance model for multicore architectures}, volume={52}, ISSN={1557-7317}, url={http://dx.doi.org/10.1145/1498765.1498785}, DOI={10.1145/1498765.1498785}, number={4}, journal={Communications of the ACM}, publisher={Association for Computing Machinery (ACM)}, author={Williams, Samuel and Waterman, Andrew and Patterson, David}, year={2009}, month=Apr, pages={65--76} }

@misc{nvidia2025rtx5090,
  author       = {{NVIDIA Corporation}},
  title        = {{NVIDIA RTX Blackwell GPU Architecture}},
  year         = {2025},
  howpublished = {\url{https://images.nvidia.com/aem-dam/Solutions/geforce/blackwell/nvidia-rtx-blackwell-gpu-architecture.pdf}},
  note         = {Accessed: 2026-10-21}
}

@misc{amd2023mi300x,
  author       = {{Advanced Micro Devices, Inc.}},
  title        = {{AMD Instinct MI300X Accelerator Data Sheet}},
  year         = {2023},
  howpublished = {\url{https://www.amd.com/content/dam/amd/en/documents/instinct-tech-docs/data-sheets/amd-instinct-mi300x-data-sheet.pdf}},
  note         = {Accessed: 2026-10-21}
}

@inproceedings{zheng2024sglang,
  title     = {{SGLang: Efficient Execution of Structured Language Model Programs}},
  author    = {Lianmin Zheng and Liangsheng Yin and Zhiqiang Xie and Chuyue Sun and Jeff Huang and Cody Hao Yu and Shiyi Cao and Christos Kozyrakis and Ion Stoica and Joseph E. Gonzalez and Clark Barrett and Ying Sheng},
  booktitle = {Advances in Neural Information Processing Systems},
  year      = {2024}
}

@inproceedings{yu2021habitat,
  title = {{Habitat: A Runtime-Based Computational Performance Predictor for Deep Neural Network Training}},
  author = {Yu, Geoffrey X. and Gao, Yubo and Golikov, Pavel and Pekhimenko, Gennady},
  booktitle = {2021 USENIX Annual Technical Conference (USENIX ATC 21)},
  year = {2021},
  url = {https://www.usenix.org/conference/atc21/presentation/yu}
}

@inproceedings{lee2025neusight,
  title = {{Forecasting GPU Performance for Deep Learning Training and Inference}},
  author = {Lee, Seonho and Phanishayee, Amar and Mahajan, Divya},
  booktitle = {Proceedings of the 30th ACM International Conference on Architectural Support for Programming Languages and Operating Systems},
  year = {2025},
  doi = {10.1145/3669940.3707265},
  url = {https://arxiv.org/abs/2407.13853}
}

@inproceedings{zhang2024llmcompass,
  title = {{LLMCompass: Enabling Efficient Hardware Design for Large Language Model Inference}},
  author = {Zhang, Hengrui and Ning, August and Prabhakar, Rohan and Wentzlaff, David},
  booktitle = {2024 ACM/IEEE 51st Annual International Symposium on Computer Architecture (ISCA)},
  year = {2024},
  pages = {1080--1096},
  doi = {10.1109/ISCA59077.2024.00082}
}

@article{li2026llmasjev,
    title   = {{LLM-as-Jev: LLMs Are Already Jev-Style Decision Models---When and How to Fine-Tune Them}},
    author  = {Yinheng Li and Justin Wagle},
    journal = {arXiv preprint arXiv:2610.02076},
    year    = {2026}
  }

@InProceedings{xiao2023smoothquant,
  title     = {{SmoothQuant}: Accurate and Efficient Post-Training
               Quantization for Large Language Models},
  author    = {Xiao, Guangxuan and Lin, Ji and Seznec, Mickael
               and Wu, Hao and Demouth, Julien and Han, Song},
  booktitle = {Proceedings of the 40th International Conference
               on Machine Learning},
  volume    = {202},
  pages     = {38087--38099},
  year      = {2023}
}

@inproceedings{he2024zipcache,
  title     = {{ZipCache}: Accurate and Efficient {KV} Cache Quantization with Salient Token Identification},
  author    = {He, Yefei and Zhang, Luoming and Wu, Weijia and Liu, Jing and Zhou, Hong and Zhuang, Bohan},
  booktitle = {Advances in Neural Information Processing Systems},
  volume    = {37},
  year      = {2024}
}

@inproceedings{li2025kvtuner,
  title     = {{KVTuner}: Sensitivity-Aware Layer-Wise Mixed-Precision {KV} Cache Quantization for Efficient and Nearly Lossless {LLM} Inference},
  author    = {Li, Xing and Xing, Zeyu and Li, Yiming and Qu, Linping and Zhen, Hui-Ling and Yao, Yiwu and Liu, Wulong and Pan, Sinno Jialin and Yuan, Mingxuan},
  booktitle = {Proceedings of the 42nd International Conference on Machine Learning},
  volume    = {267},
  pages     = {36451--36485},
  year      = {2025}
}

@misc{nvidia_rtx5090,
  author       = {{NVIDIA}},
  title        = {{GeForce RTX 5090 Specifications}},
  howpublished = {\url{https://www.nvidia.com/en-us/geforce/graphics-cards/50-series/rtx-5090/}},
  year         = {2025}
}

@misc{amd_mi300x,
  author       = {{AMD}},
  title        = {{AMD Instinct MI300X Accelerators}},
  howpublished = {\url{https://www.amd.com/en/products/accelerators/instinct/mi300/mi300x.html}},
  year         = {2023}
}

@misc{nvidia_h100,
  author       = {{NVIDIA}},
  title        = {{NVIDIA H100 Tensor Core GPU Datasheet}},
  howpublished = {\url{https://resources.nvidia.com/en-us-gpu-resources/h100-datasheet-24306}},
  year         = {2024}
}

@misc{nvidia_l40,
  author       = {{NVIDIA}},
  title        = {{NVIDIA L40 GPU Datasheet}},
  howpublished = {\url{https://www.nvidia.com/content/dam/en-zz/Solutions/design-visualization/support-guide/NVIDIA-L40-Datasheet-January-2023.pdf}},
  year         = {2023}
}
\clearpage
\appendices
\makeatletter
\setlength{\@fptop}{0pt}
\setlength{\@fpsep}{8pt}
\setlength{\@fpbot}{0pt plus 1fil}
\setlength{\@dblfptop}{0pt}
\setlength{\@dblfpsep}{8pt}
\setlength{\@dblfpbot}{0pt plus 1fil}
\makeatother

\section{\sys{} with full calibration and Results}\label{app:full calibration}
\begin{table*}[t]\centering\scriptsize
\caption{\textbf{Throughput (k tokens/s)} of decoding policies at 15.7\,GiB device budget. \textbf{Bold}: highest mean value per workload. \underline{Underlined}: the policy selected by \sys{}. Within each method (AR, vLLM SD, \sys{}), the darker shading marks fp8 KV in contrast to fp16 KV cache.}\label{tab:sel-appendix}
\setlength{\tabcolsep}{2.2pt}
\begin{tabular}{@{}r r A{4}A{10} V{10}V{22} S{8}S{18} rr@{\hspace{8pt}}A{4}A{10} V{10}V{22} S{8}S{18} rr@{}}
\toprule
 & & \multicolumn{8}{c}{H100} & \multicolumn{8}{c}{L40}\\
\cmidrule(lr){3-10}\cmidrule(l){11-18}
 & & \multicolumn{2}{c}{vLLM AR} & \multicolumn{2}{c}{vLLM SD} & \multicolumn{2}{c}{\sys{}} & \multicolumn{2}{c}{Improvement vs.} & \multicolumn{2}{c}{AR} & \multicolumn{2}{c}{vLLM SD} & \multicolumn{2}{c}{\sys{}} & \multicolumn{2}{c}{Improvement vs.}\\
\cmidrule(lr){3-4}\cmidrule(lr){5-6}\cmidrule(lr){7-8}\cmidrule(lr){9-10}\cmidrule(lr){11-12}\cmidrule(lr){13-14}\cmidrule(lr){15-16}\cmidrule(l){17-18}
$\ell$ & $B$ & fp16 & fp8 & fp16 & fp8 & fp16 & fp8 & best AR & best SD & fp16 & fp8 & fp16 & fp8 & fp16 & fp8 & best AR & best SD\\
\midrule
\multirow{4}{*}{256} & 8 & 1.20 & 1.20 & \textbf{4.19} & 4.16 & \textbf{4.19} & \underline{4.18} & $3.48\times$ & $1.00\times$ & 0.61 & 0.62 & \textbf{1.85} & 1.76 & \textbf{1.85} & \underline{1.82} & $2.94\times$ & $0.99\times$\\
 & 32 & 3.88 & 3.87 & 7.46 & 7.51 & 9.53 & \underline{\textbf{9.56}} & $2.47\times$ & $1.27\times$ & 1.97 & 2.05 & 2.16 & 2.37 & \textbf{3.65} & \underline{3.65} & $1.78\times$ & $1.54\times$\\
 & 64 & 6.90 & 6.83 & 8.33 & 8.21 & \textbf{12.57} & \underline{12.45} & $1.80\times$ & $1.49\times$ & 3.34 & 3.54 & 2.42 & 2.46 & 4.23 & \underline{\textbf{4.37}} & $1.23\times$ & $1.77\times$\\
 & 128 & 11.57 & 11.74 & 9.15 & 9.03 & \underline{\textbf{14.20}} & 13.96 & $1.21\times$ & $1.55\times$ & 5.06 & \underline{\textbf{5.44}} & 2.57 & 2.61 & 4.88 & 4.93 & $1.00\times$ & $2.09\times$\\
\cmidrule(l){1-18}
\multirow{4}{*}{1{,}024} & 8 & 1.19 & 1.20 & 3.92 & 3.91 & \underline{\textbf{3.92}} & 3.91 & $3.26\times$ & $1.00\times$ & 0.59 & 0.61 & 1.78 & 1.77 & 1.78 & \underline{\textbf{1.79}} & $2.95\times$ & $1.01\times$\\
 & 32 & 3.33 & 3.33 & 7.07 & 6.83 & \textbf{8.53} & \underline{8.22} & $2.47\times$ & $1.16\times$ & 1.73 & 1.83 & 2.17 & 2.20 & 3.43 & \underline{\textbf{3.44}} & $1.88\times$ & $1.56\times$\\
 & 64 & 5.35 & 5.40 & 7.78 & 7.48 & \textbf{10.60} & \underline{10.44} & $1.93\times$ & $1.34\times$ & 2.60 & 2.86 & 2.31 & 2.50 & 3.92 & \underline{\textbf{3.94}} & $1.38\times$ & $1.57\times$\\
 & 128 & 8.90 & 9.24 & 8.20 & 8.27 & 12.43 & \underline{\textbf{13.75}} & $1.49\times$ & $1.66\times$ & 3.96 & 4.66 & 2.28 & 2.60 & 4.00 & \underline{\textbf{4.73}} & $1.01\times$ & $1.82\times$\\
\cmidrule(l){1-18}
\multirow{4}{*}{2{,}048} & 8 & 1.16 & 1.18 & \underline{\textbf{3.84}} & 3.74 & \textbf{3.84} & 3.74 & $3.25\times$ & $1.00\times$ & 0.58 & 0.59 & 1.70 & \underline{\textbf{1.82}} & 1.70 & \textbf{1.82} & $3.06\times$ & $1.00\times$\\
 & 32 & 2.75 & 2.74 & 6.81 & 6.52 & \textbf{8.07} & \underline{7.68} & $2.80\times$ & $1.13\times$ & 1.45 & 1.55 & 2.12 & 2.22 & 3.17 & \underline{\textbf{3.33}} & $2.16\times$ & $1.50\times$\\
 & 64 & 4.14 & 4.24 & 7.35 & 6.97 & \underline{\textbf{9.03}} & 8.58 & $2.13\times$ & $1.23\times$ & 2.04 & 2.29 & 2.26 & 2.30 & \textbf{3.55} & \underline{3.55} & $1.55\times$ & $1.54\times$\\
 & 128 & 7.13 & 7.42 & 7.83 & 7.65 & 10.49 & \underline{\textbf{12.00}} & $1.62\times$ & $1.53\times$ & 3.31 & 3.67 & 2.26 & 2.55 & 3.61 & \underline{\textbf{4.51}} & $1.23\times$ & $1.77\times$\\
\midrule
\multicolumn{2}{c}{Average} & \multicolumn{6}{c}{} & $2.33\times$ & $1.28\times$ & \multicolumn{6}{c}{} & $1.85\times$ & $1.51\times$\\
\bottomrule
\end{tabular}
\end{table*}
\subsection{Cost-model initialization}
In Equation \ref{eq:mem}, the weight term and the per-token KV cost $\kappa$ follow from the model architecture. The constant overheads $\rho_g$, $C_{\text{act}}$, and $C_{\text{graph}}$ in Equation  \ref{eq:cap} depend on the GPU and software stack. For MI300X and RTX~5090, we measure them from vLLM's startup memory breakdown with $B\in\{1,32\}$, interpolating linearly in $B$. This gives $\rho_g$ of 1.14/1.25\,GiB (AR/SD) on the MI300X and 0.35/0.44\,GiB on the RTX~5090. 
We define compute factors $\eta=0.125$ on the MI300X and $\eta=0.25$ on the RTX~5090 in Eq.~\eqref{eq:ridge}. Zero-shot policy selection does not use deployment GPU startup measurements or fitted runtime overheads. 

\begin{table*}[t]\centering\scriptsize
\caption{\textbf{Speedup} of decoding policies at 15.7\,GiB device budget. \textbf{Bold}: highest mean value per workload. \underline{Underlined}: the policy selected by \sys{}. Within each method (AR, vLLM SD, \sys{}), the darker shading marks fp8 KV in contrast to fp16 KV cache.}\label{tab:speedup-appendix}
\setlength{\tabcolsep}{2.2pt}
\begin{tabular}{@{}r r A{4}A{10} V{10}V{22} S{8}S{18} rr@{\hspace{8pt}}A{4}A{10} V{10}V{22} S{8}S{18} rr@{}}
\toprule
 & & \multicolumn{8}{c}{H100} & \multicolumn{8}{c}{L40}\\
\cmidrule(lr){3-10}\cmidrule(l){11-18}
 & & \multicolumn{2}{c}{vLLM AR} & \multicolumn{2}{c}{vLLM SD} & \multicolumn{2}{c}{\sys{}} & \multicolumn{2}{c}{Improvement vs.} & \multicolumn{2}{c}{AR} & \multicolumn{2}{c}{Baseline SD} & \multicolumn{2}{c}{\sys{}} & \multicolumn{2}{c}{Improvement vs.}\\
\cmidrule(lr){3-4}\cmidrule(lr){5-6}\cmidrule(lr){7-8}\cmidrule(lr){9-10}\cmidrule(lr){11-12}\cmidrule(lr){13-14}\cmidrule(lr){15-16}\cmidrule(l){17-18}
$\ell$ & $B$ & fp16 & fp8 & fp16 & fp8 & fp16 & fp8 & best AR & best SD & fp16 & fp8 & fp16 & fp8 & fp16 & fp8 & best AR & best SD\\
\midrule
\multirow{4}{*}{256} & 8 & 1.00$\times$ & 1.00$\times$ & 3.49$\times$ & 3.48$\times$ & 3.49$\times$ & \underline{\textbf{3.49$\times$}} & $3.49\times$ & $1.00\times$ & 1.00$\times$ & 1.01$\times$ & \textbf{3.02$\times$} & 2.87$\times$ & \textbf{3.02$\times$} & \underline{2.97$\times$} & $2.94\times$ & $0.98\times$\\
 & 32 & 1.00$\times$ & 1.00$\times$ & 1.92$\times$ & 1.94$\times$ & 2.46$\times$ & \underline{\textbf{2.47$\times$}} & $2.47\times$ & $1.27\times$ & 1.00$\times$ & 1.04$\times$ & 1.09$\times$ & 1.20$\times$ & \textbf{1.85$\times$} & \underline{1.85$\times$} & $1.78\times$ & $1.54\times$\\
 & 64 & 1.00$\times$ & 0.99$\times$ & 1.21$\times$ & 1.19$\times$ & \textbf{1.83$\times$} & \underline{1.81$\times$} & $1.81\times$ & $1.49\times$ & 1.00$\times$ & 1.06$\times$ & 0.73$\times$ & 0.74$\times$ & 1.27$\times$ & \underline{\textbf{1.31$\times$}} & $1.23\times$ & $1.78\times$\\
 & 128 & 1.00$\times$ & 1.01$\times$ & 0.79$\times$ & 0.78$\times$ & \underline{\textbf{1.23$\times$}} & 1.21$\times$ & $1.21\times$ & $1.55\times$ & 1.00$\times$ & \underline{\textbf{1.08$\times$}} & 0.51$\times$ & 0.51$\times$ & 0.97$\times$ & 0.98$\times$ & $1.00\times$ & $2.09\times$\\
\cmidrule(l){1-18}
\multirow{4}{*}{1{,}024} & 8 & 1.00$\times$ & 1.03$\times$ & \textbf{3.33$\times$} & 3.27$\times$ & \underline{3.31$\times$} & 3.27$\times$ & $3.23\times$ & $1.00\times$ & 1.00$\times$ & 1.03$\times$ & 2.97$\times$ & 2.95$\times$ & 2.97$\times$ & \underline{\textbf{2.99$\times$}} & $2.91\times$ & $1.01\times$\\
 & 32 & 1.00$\times$ & 0.99$\times$ & 2.09$\times$ & 2.05$\times$ & \textbf{2.51$\times$} & \underline{2.46$\times$} & $2.46\times$ & $1.18\times$ & 1.00$\times$ & 1.06$\times$ & 1.27$\times$ & 1.28$\times$ & 1.99$\times$ & \underline{\textbf{2.01$\times$}} & $1.90\times$ & $1.57\times$\\
 & 64 & 1.00$\times$ & 1.01$\times$ & 1.44$\times$ & 1.39$\times$ & \textbf{1.97$\times$} & \underline{1.94$\times$} & $1.92\times$ & $1.34\times$ & 1.00$\times$ & 1.08$\times$ & 0.88$\times$ & 0.95$\times$ & 1.50$\times$ & \underline{\textbf{1.51$\times$}} & $1.40\times$ & $1.58\times$\\
 & 128 & 1.00$\times$ & 1.04$\times$ & 0.93$\times$ & 0.93$\times$ & 1.40$\times$ & \underline{\textbf{1.54$\times$}} & $1.49\times$ & $1.66\times$ & 1.00$\times$ & 1.18$\times$ & 0.58$\times$ & 0.66$\times$ & 1.01$\times$ & \underline{\textbf{1.20$\times$}} & $1.01\times$ & $1.82\times$\\
\cmidrule(l){1-18}
\multirow{4}{*}{2{,}048} & 8 & 1.00$\times$ & 1.02$\times$ & \textbf{3.32$\times$} & 3.16$\times$ & \underline{\textbf{3.32$\times$}} & 3.16$\times$ & $3.26\times$ & $1.00\times$ & 1.00$\times$ & 1.05$\times$ & 2.97$\times$ & \textbf{3.13$\times$} & 2.97$\times$ & \underline{\textbf{3.13$\times$}} & $2.99\times$ & $1.00\times$\\
 & 32 & 1.00$\times$ & 0.98$\times$ & 2.46$\times$ & 2.37$\times$ & \textbf{2.91$\times$} & \underline{2.79$\times$} & $2.79\times$ & $1.13\times$ & 1.00$\times$ & 1.06$\times$ & 1.47$\times$ & 1.51$\times$ & 2.18$\times$ & \underline{\textbf{2.23$\times$}} & $2.11\times$ & $1.48\times$\\
 & 64 & 1.00$\times$ & 1.00$\times$ & 1.75$\times$ & 1.67$\times$ & \underline{\textbf{2.15$\times$}} & 2.04$\times$ & $2.15\times$ & $1.23\times$ & 1.00$\times$ & 1.12$\times$ & 1.13$\times$ & 1.13$\times$ & 1.75$\times$ & \underline{\textbf{1.76$\times$}} & $1.57\times$ & $1.56\times$\\
 & 128 & 1.00$\times$ & 1.02$\times$ & 1.08$\times$ & 1.04$\times$ & 1.45$\times$ & \underline{\textbf{1.64$\times$}} & $1.60\times$ & $1.51\times$ & 1.00$\times$ & 1.09$\times$ & 0.68$\times$ & 0.76$\times$ & 1.09$\times$ & \underline{\textbf{1.36$\times$}} & $1.25\times$ & $1.79\times$\\
\midrule
\multicolumn{2}{c}{Average} & \multicolumn{6}{c}{} & $2.32\times$ & $1.28\times$ & \multicolumn{6}{c}{} & $1.84\times$ & $1.52\times$\\
\bottomrule
\end{tabular}
\end{table*}

\begin{table*}[t]
\centering
\scriptsize
\caption{MI300X \textbf{throughput (k tokens/s)} for every static draft length and KV-cache precision at the 15.7\,GiB budget. \textbf{Bold}: highest mean throughput at that workload.}
\label{tab:fixedk-throughput}
\setlength{\tabcolsep}{5pt}
\begin{tabular}{@{}rr rr rr rr@{}}
\toprule
 & & \multicolumn{2}{c}{$k=3$} & \multicolumn{2}{c}{$k=7$} & \multicolumn{2}{c}{$k=15$}\\
\cmidrule(lr){3-4}\cmidrule(lr){5-6}\cmidrule(lr){7-8}
$\ell$ & $B$ & fp16 & fp8 & fp16 & fp8 & fp16 & fp8\\
\midrule
\multirow{4}{*}{256} & 8 & 3.49 & 3.75 & 4.32 & \textbf{4.63} & 4.26 & 4.55\\
 & 32 & 8.15 & 8.90 & 8.35 & \textbf{9.12} & 7.18 & 8.14\\
 & 64 & 10.69 & \textbf{11.30} & 10.09 & 11.28 & 8.45 & 8.93\\
 & 128 & 13.15 & \textbf{14.21} & 12.35 & 12.53 & 9.38 & 10.13\\
\cmidrule(lr){1-8}
\multirow{4}{*}{1{,}024} & 8 & 2.94 & 3.47 & 3.68 & \textbf{4.31} & 3.58 & 4.18\\
 & 32 & 7.01 & \textbf{8.15} & 7.03 & 8.13 & 6.09 & 6.89\\
 & 64 & 8.46 & 9.65 & 8.55 & \textbf{9.89} & 7.21 & 8.53\\
 & 128 & 9.54 & \textbf{11.92} & 9.36 & 10.75 & 8.14 & 10.13\\
\cmidrule(lr){1-8}
\multirow{4}{*}{2{,}048} & 8 & 2.64 & 3.19 & 3.38 & 4.01 & 3.42 & \textbf{4.03}\\
 & 32 & 5.80 & 6.70 & 6.39 & \textbf{6.79} & 5.50 & 6.47\\
 & 64 & 6.88 & \textbf{8.20} & 6.86 & 7.76 & 6.39 & 7.48\\
 & 128 & 7.62 & \textbf{9.04} & 7.71 & 8.58 & 6.90 & 8.91\\
\bottomrule
\end{tabular}
\end{table*}

\begin{table*}[t]
\centering
\fontsize{6}{6.5}\selectfont
\renewcommand{\arraystretch}{0.9}
\caption{MI300X \textbf{wall-clock speedup} over AR decoding with fp16 KV for every static draft length and KV-cache precision. \textbf{Bold}: highest mean speedup at that workload.}
\label{tab:fixedk-speedup}
\setlength{\tabcolsep}{5pt}
\begin{tabular}{@{}rr rr rr rr@{}}
\toprule
 & & \multicolumn{2}{c}{$k=3$} & \multicolumn{2}{c}{$k=7$} & \multicolumn{2}{c}{$k=15$}\\
\cmidrule(lr){3-4}\cmidrule(lr){5-6}\cmidrule(lr){7-8}
$\ell$ & $B$ & fp16 & fp8 & fp16 & fp8 & fp16 & fp8\\
\midrule
\multirow{4}{*}{256} & 8 & $1.86\times$ & $2.01\times$ & $2.31\times$ & \textbf{2.47$\times$} & $2.27\times$ & $2.42\times$\\
 & 32 & $1.52\times$ & $1.67\times$ & $1.56\times$ & \textbf{1.70$\times$} & $1.39\times$ & $1.58\times$\\
 & 64 & $1.34\times$ & \textbf{1.41$\times$} & $1.26\times$ & \textbf{1.41$\times$} & $1.03\times$ & $1.08\times$\\
 & 128 & $1.21\times$ & \textbf{1.31$\times$} & $1.13\times$ & $1.15\times$ & $0.78\times$ & $0.85\times$\\
\cmidrule(lr){1-8}
\multirow{4}{*}{1{,}024} & 8 & $1.64\times$ & $1.93\times$ & $2.05\times$ & \textbf{2.41$\times$} & $1.99\times$ & $2.34\times$\\
 & 32 & $1.72\times$ & \textbf{2.02$\times$} & $1.72\times$ & $1.97\times$ & $1.57\times$ & $1.76\times$\\
 & 64 & $1.42\times$ & $1.60\times$ & $1.43\times$ & \textbf{1.64$\times$} & $1.22\times$ & $1.44\times$\\
 & 128 & $1.27\times$ & \textbf{1.57$\times$} & $1.25\times$ & $1.39\times$ & $0.98\times$ & $1.22\times$\\
\cmidrule(lr){1-8}
\multirow{4}{*}{2{,}048} & 8 & $1.44\times$ & $1.72\times$ & $1.89\times$ & $2.21\times$ & $1.98\times$ & \textbf{2.32$\times$}\\
 & 32 & $1.80\times$ & $1.95\times$ & $1.91\times$ & $1.98\times$ & $1.77\times$ & \textbf{2.01$\times$}\\
 & 64 & $1.39\times$ & \textbf{1.62$\times$} & $1.40\times$ & $1.53\times$ & $1.38\times$ & $1.60\times$\\
 & 128 & $1.30\times$ & \textbf{1.52$\times$} & $1.34\times$ & $1.40\times$ & $1.13\times$ & $1.46\times$\\
\bottomrule
\end{tabular}
\end{table*}
\section{\sys{} with Learning-Based Policy Selectors}\label{app:transfer-training}
\textbf{Data and features.} Each deployment GPU is excluded from the three-GPU source dataset. Each source contributes 96 configuration--workload points: eight static choices at 12 workloads, with throughput aggregated over the available measurement seeds. For IDW, the shallow tree, and TabICL, numerical deployment metadata features encode memory budget, physical memory capacity, submitted request count, and whether the software platform uses ROCm. Workload features encode normalized log concurrency and output cap. Candidate features encode AR versus SD, draft length, and KV precision. The compact model uses only workload and candidate features. Jev receives these settings as text, together with the recorded attention backend, weight precision, and software versions where available.

\textbf{IDW and shallow tree.} IDW predicts log-throughput with weights $w_i=(d_i^2+0.01)^{-2}$, where $d_i^2$ is squared Euclidean distance in the fixed feature representation. It fits no model parameters. The tree uses maximum depth 2 and random seed 0. Training labels are the fastest observed configurations grouped by source GPU and workload.

\textbf{TabICL.} We use the pretrained TabICL regressor with eight estimators, GPU inference, and the trial seed as its random seed. Inputs are the numerical features above and labels are log-throughput. Source measurements condition its predictions. No gradient fine-tuning of the pretrained model is performed.

\textbf{Compact numerical model.} Separate workload and candidate encoders feed a shared scoring head, with hidden width 32 and GELU activations. Training minimizes centered log-throughput mean squared error plus $0.1$ times a within-workload soft-ranking cross-entropy term, with temperature $0.1$. AdamW uses learning rate $0.003$, weight decay $0.01$, and seed 0. Source-only validation leaves out each source GPU in turn and compares 300 versus 900 training steps and ridge-correction penalties $\alpha\in\{0.1,1,10\}$, using three random 16-point supports on each validation GPU. The final model is refit on all three sources. All four folds select 900 steps. The source-validation correction penalty is 0.1 for RTX~5090 and 1 for the other targets. The zero-shot evaluation uses the source-trained model without a deployment-GPU residual correction.

\textbf{Jev decision model.} Both variants use Qwen3-4B with an eight-choice answer readout. Prompts contain the workload and the deployment settings listed above, together with up to 16 retrieved measurements per source GPU, selected by workload proximity without ranking by throughput. The source-tuned variant trains rank-8 LoRA adapters on attention query, key, value, and output projections ($\alpha=16$, dropout 0.05), together with a zero-initialized linear correction to the answer logits. Training uses classification cross-entropy, AdamW with learning rate $10^{-4}$ and weight decay 0.01, effective batch size 8, gradient clipping at 1, and seed 0. Execution uses FP32 with SDPA and an 8,192-token prompt limit.

During training on source GPUs, prediction tasks simulate 0, 4, 8, or 16 observations from a source GPU treated as the query device, excluding query-workload outcomes from prediction inputs. These observations come entirely from the source dataset. Evaluation on the deployment GPU uses zero-shot policy selection. One supplied-value reading task accompanies every four prediction tasks; answer order is shuffled consistently with the labels. One of the three source GPUs is reserved for development validation, with training on the other two. Checkpoints at 0, 25, 50, 100, and 200 steps are compared by validation throughput loss, then the chosen duration is used to refit from initialization on all three sources. The selected durations are 100 steps for RTX~5090 and MI300X and 200 for H100 and L40. The frozen variant retains the initial base model and zero correction. Supplied-value reading checks are recorded separately and are not uniformly perfect.

\textbf{Selected policies.}\label{app:transfer-maps}
Table~\ref{tab:transfer-policies} reports the saved static choices from zero-shot policy selection for IDW and source-tuned Jev in the format of Table~\ref{tab:selpol}. Table~\ref{tab:transfer-maps} gives dynamic mappings constructed from their saved candidate scores, using the same batch ranges as Table~\ref{tab:maps}. For each GPU, output cap, KV precision, and concurrency anchor $B\in\{8,32,64,128\}$, we maximize the score over $k\in\{3,7,15\}$; exact ties prefer smaller $k$. The anchors define ranges 1--20, 21--48, 49--96, and 97--128. AR is excluded from this within-engine mapping; choosing AR or a KV precision remains a separate startup decision.

IDW uses predicted log-throughput; Jev uses the eight-choice answer logits restricted to the three speculative actions at the specified KV precision. Neither construction uses target outcomes. These tables characterize the dynamic mappings implied by the static candidate scores. The performance evaluation is restricted to static configuration selection; it does not measure the throughput of these dynamic mappings.

\begin{table*}[t]\centering\fontsize{6}{6.5}\selectfont
\renewcommand{\arraystretch}{0.9}
\caption{Policies selected by IDW and source-tuned Jev at each workload: AR vs. vLLM SD and KV precision.}
\label{tab:transfer-policies}
\setlength{\tabcolsep}{7pt}
\begin{tabular}{@{}llrcccc@{}}
\toprule
Method & GPU & $\ell$ & $B{=}8$ & $B{=}32$ & $B{=}64$ & $B{=}128$\\
\midrule
IDW & MI300X & 256 & $k{=}7$, fp8 & $k{=}7$, fp8 & $k{=}7$, fp8 & $k{=}3$, fp8\\
 &  & 1{,}024 & $k{=}7$, fp8 & $k{=}7$, fp8 & $k{=}7$, fp8 & $k{=}3$, fp8\\
 &  & 2{,}048 & $k{=}7$, fp8 & $k{=}7$, fp8 & $k{=}7$, fp8 & $k{=}3$, fp8\\
\midrule
IDW & RTX~5090 & 256 & $k{=}15$, fp16 & $k{=}3$, fp8 & $k{=}3$, fp8 & AR, fp8\\
 &  & 1{,}024 & $k{=}15$, fp8 & $k{=}3$, fp8 & $k{=}3$, fp8 & $k{=}3$, fp8\\
 &  & 2{,}048 & $k{=}15$, fp8 & $k{=}3$, fp8 & $k{=}3$, fp8 & $k{=}3$, fp8\\
\midrule
IDW & H100 & 256 & $k{=}15$, fp16 & $k{=}3$, fp8 & $k{=}3$, fp8 & $k{=}3$, fp8\\
 &  & 1{,}024 & $k{=}15$, fp8 & $k{=}3$, fp8 & $k{=}3$, fp8 & $k{=}3$, fp8\\
 &  & 2{,}048 & $k{=}15$, fp8 & $k{=}3$, fp8 & $k{=}3$, fp8 & $k{=}3$, fp8\\
\midrule
IDW & L40 & 256 & $k{=}15$, fp16 & $k{=}3$, fp8 & $k{=}3$, fp8 & $k{=}3$, fp8\\
 &  & 1{,}024 & $k{=}15$, fp16 & $k{=}3$, fp8 & $k{=}3$, fp8 & $k{=}3$, fp8\\
 &  & 2{,}048 & $k{=}15$, fp8 & $k{=}7$, fp8 & $k{=}3$, fp8 & $k{=}3$, fp8\\
\midrule
Jev & MI300X & 256 & $k{=}15$, fp16 & $k{=}3$, fp8 & $k{=}3$, fp8 & $k{=}3$, fp8\\
 &  & 1{,}024 & $k{=}15$, fp16 & $k{=}3$, fp8 & $k{=}3$, fp8 & $k{=}3$, fp8\\
 &  & 2{,}048 & $k{=}15$, fp8 & $k{=}7$, fp8 & $k{=}3$, fp8 & $k{=}3$, fp8\\
\midrule
Jev & RTX~5090 & 256 & $k{=}15$, fp8 & $k{=}3$, fp8 & $k{=}3$, fp8 & AR, fp8\\
 &  & 1{,}024 & $k{=}15$, fp16 & $k{=}3$, fp8 & $k{=}3$, fp8 & $k{=}3$, fp8\\
 &  & 2{,}048 & $k{=}15$, fp8 & $k{=}3$, fp8 & $k{=}3$, fp8 & $k{=}3$, fp8\\
\midrule
Jev & H100 & 256 & $k{=}15$, fp16 & $k{=}3$, fp8 & $k{=}3$, fp8 & $k{=}3$, fp8\\
 &  & 1{,}024 & $k{=}15$, fp8 & $k{=}3$, fp8 & $k{=}3$, fp8 & $k{=}3$, fp8\\
 &  & 2{,}048 & $k{=}15$, fp8 & $k{=}3$, fp8 & $k{=}3$, fp8 & $k{=}3$, fp8\\
\midrule
Jev & L40 & 256 & $k{=}15$, fp16 & $k{=}3$, fp8 & $k{=}3$, fp8 & $k{=}3$, fp8\\
 &  & 1{,}024 & $k{=}15$, fp8 & $k{=}3$, fp8 & $k{=}3$, fp8 & $k{=}3$, fp8\\
 &  & 2{,}048 & $k{=}15$, fp8 & $k{=}7$, fp8 & $k{=}3$, fp8 & $k{=}3$, fp8\\
\bottomrule
\end{tabular}
\end{table*}

\begin{table*}[t]\centering\fontsize{6}{6.5}\selectfont
\renewcommand{\arraystretch}{0.9}
\caption{Predicted dynamic draft-length mappings of IDW and source-tuned Jev. Each entry reports $k$ for fp16/fp8 KV, respectively, at the indicated effective-batch range.}
\label{tab:transfer-maps}
\setlength{\tabcolsep}{7pt}
\begin{tabular}{@{}llrcccc@{}}
\toprule
Method & GPU & $\ell$ & 1--20 & 21--48 & 49--96 & 97--128\\
\midrule
IDW & MI300X & 256 & 7/7 & 7/7 & 7/7 & 3/3\\
 &  & 1{,}024 & 7/7 & 7/7 & 7/7 & 3/3\\
 &  & 2{,}048 & 7/7 & 7/7 & 7/7 & 3/3\\
\midrule
IDW & RTX~5090 & 256 & 15/15 & 3/3 & 3/3 & 3/3\\
 &  & 1{,}024 & 15/15 & 3/3 & 3/3 & 3/3\\
 &  & 2{,}048 & 15/15 & 3/3 & 3/3 & 3/3\\
\midrule
IDW & H100 & 256 & 15/15 & 3/3 & 3/3 & 3/3\\
 &  & 1{,}024 & 15/15 & 3/3 & 3/3 & 3/3\\
 &  & 2{,}048 & 15/15 & 3/3 & 3/3 & 3/3\\
\midrule
IDW & L40 & 256 & 15/15 & 3/3 & 3/3 & 3/3\\
 &  & 1{,}024 & 15/15 & 7/3 & 3/3 & 3/3\\
 &  & 2{,}048 & 15/15 & 7/7 & 3/3 & 3/3\\
\midrule
Jev & MI300X & 256 & 15/15 & 3/3 & 3/3 & 3/3\\
 &  & 1{,}024 & 15/15 & 3/3 & 3/3 & 3/3\\
 &  & 2{,}048 & 15/15 & 3/7 & 3/3 & 3/3\\
\midrule
Jev & RTX~5090 & 256 & 15/15 & 3/3 & 3/3 & 3/3\\
 &  & 1{,}024 & 15/15 & 3/3 & 3/3 & 3/3\\
 &  & 2{,}048 & 15/15 & 3/3 & 3/3 & 3/3\\
\midrule
Jev & H100 & 256 & 15/15 & 3/3 & 3/3 & 3/3\\
 &  & 1{,}024 & 15/15 & 3/3 & 3/3 & 3/3\\
 &  & 2{,}048 & 15/15 & 3/3 & 3/3 & 3/3\\
\midrule
Jev & L40 & 256 & 15/15 & 3/3 & 3/3 & 3/3\\
 &  & 1{,}024 & 15/15 & 3/3 & 3/3 & 3/3\\
 &  & 2{,}048 & 15/15 & 3/7 & 3/3 & 3/3\\
\bottomrule
\end{tabular}
\end{table*}

\end{document}